\pdfoutput=1
\documentclass{article}
\usepackage{iclr2027_preprint,times}

\usepackage[hyphens]{url}
\usepackage{graphicx}
\usepackage{algorithm}
\usepackage{algpseudocode}
\usepackage{booktabs}
\usepackage{amsmath,amssymb,amsthm,mathtools,bm}
\usepackage{xcolor}
\usepackage{tcolorbox}
\tcbuselibrary{breakable}
\usepackage{enumitem}
\usepackage{array}
\usepackage{multirow}
\usepackage{tabularx}
\newcommand{\R}{\mathbb{R}}

\newcommand{\Stiefel}{\mathrm{St}}

\newcommand{\tr}{\mathrm{tr}}
\newcommand{\logdet}{\log\det}

\newcommand{\polar}{\mathrm{Polar}}
\newcommand{\rank}{\mathrm{rank}}
\newcommand{\diag}{\mathrm{diag}}
\newcommand{\T}{^{\!\top}}
\newcommand{\NEPv}{\textnormal{\textsc{NEPv}}}
\newcommand{\DPP}{\textnormal{\textsc{DPP}}}
\newcommand{\SCF}{\textnormal{\textsc{SCF}}}

\newcommand{\SoftMin}{\operatorname{softmin}}
\newcommand{\LO}{\mathrm{LO}}
\definecolor{cBlue}{HTML}{1f77b4}
\definecolor{cOrange}{HTML}{ff7f0e}
\definecolor{cGreen}{HTML}{2ca02c}
\definecolor{cRed}{HTML}{d62728}
\definecolor{cPurple}{HTML}{9467bd}
\definecolor{cGray}{HTML}{7f7f7f}
\definecolor{cTeal}{HTML}{17becf}
\newtheorem{proposition}{Proposition}
\newtheorem{theorem}{Theorem}
\newtheorem{lemma}{Lemma}

\theoremstyle{definition}
\newtheorem{definition}{Definition}
\newtheorem{assumption}{Assumption}
\theoremstyle{remark}
\newtheorem{remark}{Remark}
\newtheorem*{example*}{Example}
\usepackage[colorlinks=true,linkcolor=cBlue,citecolor=cGreen,urlcolor=cTeal,
            breaklinks=true]{hyperref}
\hypersetup{
  pdftitle={Fair Subset Selection across Overlapping Categories},
  pdfauthor={Richard Yi Da Xu, Junhong Zhang, and Alan Nguyen}
}
\usepackage{cleveref}
\crefname{assumption}{Assumption}{Assumptions}
\Crefname{assumption}{Assumption}{Assumptions}

\title{Fair Subset Selection across\\Overlapping Categories}

\author{Richard Yi Da Xu\textsuperscript{1}, Junhong Zhang\textsuperscript{2}, and Alan Nguyen\textsuperscript{3} \\
\textsuperscript{1}Hong Kong Baptist University \quad
\textsuperscript{2}Shenzhen University \quad
\textsuperscript{3}TadReamk Limited \\
{\fontsize{8.2}{10}\selectfont\texttt{xuyida@hkbu.edu.hk}, \texttt{zhangjunhong2022@email.szu.edu.cn}, \texttt{alannguyen@tadreamk.com}}
}

\iclrfinalcopy

\begin{document}
\maketitle
\begin{abstract}
Selecting a diverse subset across overlapping categories allocates one budget
to items that can serve several categories at once. We score each category by
a log-determinant with a membership floor and ridge, then optimize a soft
minimum of those scores. We ask when the extra cost of a Stiefel relaxation
solved by adaptive self-consistent field (SCF) earns a better discrete subset
than greedy selection. The two approaches differ most when near-duplicate
items cover several weak categories. We construct instances on which greedy
followed by one-item swaps incurs an unbounded loss under the exact minimum,
and test the pattern on naturally redundant VisDrone frames. We derive a
level-shift condition for monotone ascent of the relaxed objective and compare
SCF with locally optimal conjugate gradient (LOCG) acceleration inside its
nonlinear iteration. On near-duplicate VisDrone frames the relaxation beats a multistart
baseline in all 11 paired comparisons at budget 20. The gains fade at
nearby budgets. On COCO, LOCG lowers the final stationarity residual,
while greedy gives the best discrete score and probe mAP. A safeguard returns the higher-scoring of the greedy and relaxed
subsets and therefore matches or improves the exact discrete objective
of greedy on every instance. Greedy remains substantially cheaper and
improves downstream accuracy over uniform sampling on long-tailed
classification benchmarks.
\end{abstract}
\section{Introduction}\label{sec:intro}
\begin{figure}[!b]
\centering
\includegraphics[width=\linewidth]{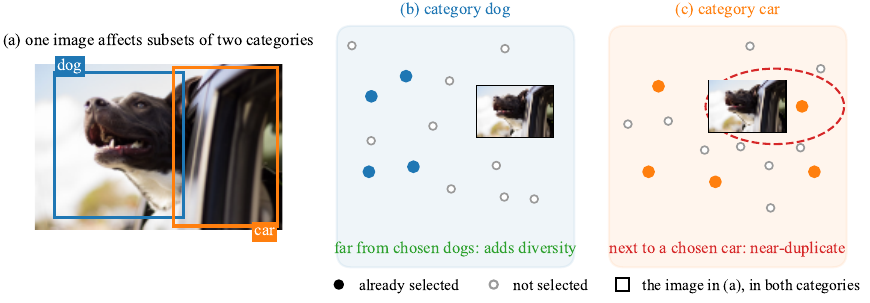}
\caption{\textbf{One image, two categories.} (a)~An image showing a dog and a
car takes one of the $k$ places in the subset but belongs to both categories.
(b,c)~Dots are items of each category in feature space. The image lies far
from the dogs already selected, so it adds diversity to \emph{dog}, but next
to a car already selected, so it is a near-duplicate for \emph{car}.}
\label{fig:intro-example}
\end{figure}

A dataset can contain many more items than a labeling or training budget
allows. The challenge in selecting a subset is to represent rare categories
while avoiding repetition. In multi-label data the choices are coupled.
One photo of both a dog and a car uses one place and counts toward both
categories, even if its dog is novel and its car is nearly a duplicate
(\cref{fig:intro-example}). Categories therefore compete for a shared budget
and benefit from some of the same items.

A single feature Gram determinant rewards diversity across the subset
\citep{kulesza2012dpp}. Group quotas protect representation by fixing
counts in advance \citep{celis2018fair}. To model the interaction directly,
we evaluate the same subset with one kernel per category.
A membership floor downweights nonmembers and a ridge keeps all category
scores finite. The soft minimum of the category log-determinants emphasizes
the weakest categories. We use fairness to refer to this objective.
The floor and ridge determine how strongly the objective rewards member
counts relative to feature diversity (\cref{sec:model,sec:fair-objective}).

Greedy selection is attractive because a shared Cholesky factorization
gives linear scaling in the number of candidates for fixed feature width,
budget, and category count. That same one-item update can become a
liability when several weak categories share near-duplicate items.
A duplicate can help all of their counts at once, and replacing it may
require coordinated changes. The Stiefel relaxation can move all budget
directions together. Its usefulness depends on whether rounding converts
that larger search space into a better subset at an acceptable cost.
This raises two questions. When does the relaxation beat greedy and
exchange baselines? When does improving the continuous solve help
the final subset?

\newpage
\paragraph{Contributions.}
\begin{enumerate}[leftmargin=*]
\item \textbf{Category objectives and greedy limits.} The category kernels
account for overlapping membership under a common budget. We show when greedy
attains the largest possible minimum count on partitions and construct
near-duplicate, shared-label instances where exact-minimum greedy with
one-item swaps has unbounded loss as the floor shrinks
(\cref{sec:model,sec:fair-objective,sec:greedy}).
\item \textbf{Adaptive SCF and ascent.} The determinant-specific lifting has
an explicit spectrum and an exact eigensolve of size at most $2k$.
We establish a sufficient level shift for monotone ascent and test LOCG,
which applies several nonlinear SCF sweeps in a subspace of dimension
at most $3k$ (\cref{sec:solver-comparison,supp:sec:locg}).
\item \textbf{When relaxation pays off.} Paired VisDrone and COCO experiments
compare stationarity, selection time, and rounded quality against greedy.
VisDrone tests the shared-label, near-duplicate regime and nearby budgets.
COCO exposes the gap between relaxed and discrete quality. A safeguarded
selector preserves the baseline's discrete score. Long-tailed CIFAR and
GTSRB evaluate the cheaper greedy selector against sampling
and diversity baselines (\cref{sec:solver-comparison,sec:experiments-real}).
\end{enumerate}

\section{Background and Related Work}\label{sec:related}\label{sec:bg}
For self-containedness, we first review the determinantal background our method builds on, and
then position it among related work on fair selection and data selection.

\paragraph{Determinantal selection.}
For a positive-semidefinite matrix $L$ on $[n]=\{1,\ldots,n\}$, an
$L$-ensemble \DPP\ assigns probability $\det(L_S)/\det(I+L)$ to a subset $S$,
where $L_S$ is the principal submatrix indexed by $S$; it models quality and
diversity through these determinants \citep{kulesza2012dpp}. Conditioning on
size $k$ gives the MAP problem $\max_{|S|=k}\logdet(L_S)$, with $\log0=-\infty$,
which is NP-hard and closely
related to maximum-entropy sampling and maximum-volume subset selection
\citep{ko1995entropy,civril2009selecting}. Greedy and accelerated
log-determinant methods select items directly \citep{han2017faster,chen2018fastdpp},
and fractional relaxations optimize item weights
\citep{gillenwater2012near,nikolov2015randomized,singh2020doptimal}. Our
relaxation instead optimizes over orthonormal frames, which contain every
coordinate selector; orthogonality-constrained determinant objectives also
appear in PCA \citep{hauser2018determinant}. Our rounding builds on leverage-score
and exchange methods \citep{gu1996rrqr,belhadji2020dppcss}.

\paragraph{Representation and robust aggregation.}
Fair \DPP s and partition-constrained determinant maximization enforce
prescribed counts per group \citep{celis2018fair,nikolov2016partition}. We
instead encode possibly overlapping memberships in category kernels and raise
the weakest category diversity without fixing counts. How the categories are
aggregated matters: a fixed-weight product of determinants, equivalently a sum
of log-determinants \citep{ohsaka2020product}, lets strong categories
compensate for weak ones, and averaging kernels before taking a determinant
loses per-category control. Multiple-kernel \DPP s learn or mix kernels
\citep{chao2015largemargin,matsuoka2021mixtures} rather than scoring one
subset against every category. Maximin design, robust subset selection, including the SATURATE
algorithm for maximizing a minimum of submodular functions
\citep{dette2007maximin,krause2008robust,udwani2018multiobjective}, and soft-min
smoothing \citep{nesterov2005smooth} are established; we apply them to category
determinants, where robust-submodular guarantees do not directly carry over
because log-determinants need not be monotone. Fair PCA balances group
utilities with variance criteria \citep{samadi2018fairpca,tantipongpipat2019multicriteria}.

\paragraph{Training-data selection and downstream tasks.}
Determinantal methods also build coresets, with guarantees on approximating a
downstream loss \citep{tremblay2019coresets} or the determinant maximum itself
\citep{gollapudi2023composable}, and small curated training sets can match
much larger ones in instruction tuning \citep{zhou2023lima,chen2024alpagasus}.
We share their aim of training good models from few examples, but tackle a
different difficulty: keeping every category represented and diverse when one
item can belong to several categories. Like them, we judge a subset by how well a
model trained on it performs (\cref{sec:experiments-real}).

\section{Category Kernels}\label{sec:model}\label{sec:group-selection}\label{sec:independent-selection}\label{sec:group-kernels}
\begin{figure}[!t]
\centering
\includegraphics[width=0.85\linewidth]{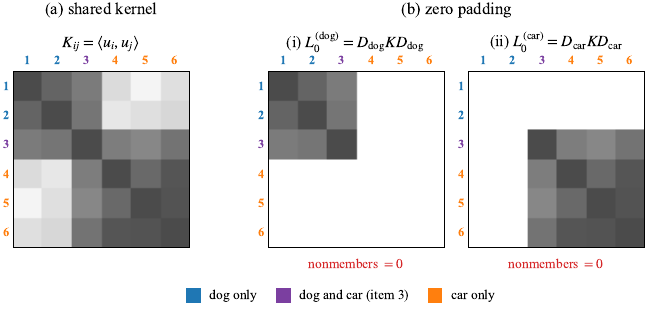}
\caption{\textbf{Category kernels before softening.} Items 1--3 are dogs and
items 3--6 are cars; item 3 belongs to both. (a)~The shared kernel $K$.
(b)~Zero padding: (i)~$L_0^{(\mathrm{dog})}$ keeps only the dog block of $K$, so any
subset containing a car-only item has determinant zero for dog; (ii)~$L_0^{(\mathrm{car})}$
keeps only the car block. Item 3 appears in both.}
\label{fig:kernel-base}
\end{figure}

\begin{figure}[!t]
\centering
\includegraphics[width=\linewidth]{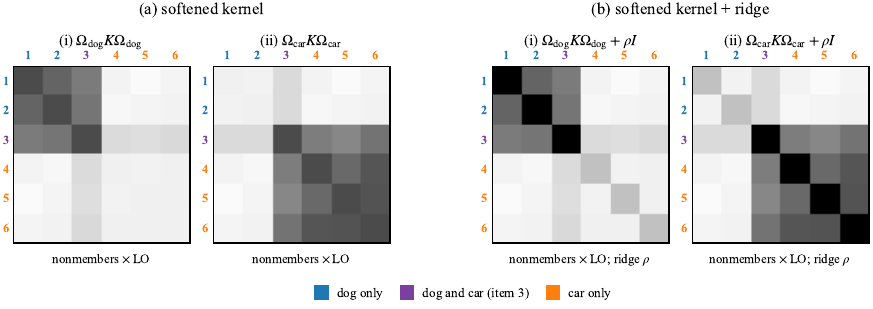}
\caption{\textbf{Softened category kernels} (same items as
\cref{fig:kernel-base}), (a)~without and (b)~with the ridge $\rho I$, for (i)~dog and (ii)~car. Each
category keeps its own block at full weight and fades nonmembers by $\LO$; the
ridge gives every nonmember at least $\rho$ on the diagonal, so every subset
receives a finite score. Item 3 lies in both blocks, so it receives full
membership weight in both categories. ($\LO$ and $\rho$ are enlarged for visibility.)}
\label{fig:kernel-soft}
\end{figure}

As stated in the introduction, a key design element of our method is the kernel
through which each category scores a subset; we call these \emph{category kernels}
(Figures~\ref{fig:kernel-base}(b) and~\ref{fig:kernel-soft}). Below we follow the
journey of our kernel design: we begin with the kernel design, move from the shared
kernel, through zero padding, to our softened category kernels, and end with how
they trade a category's own diversity against its member count and why they favor
items shared by several categories.

\paragraph{Kernel design.}
Each item $i$ has a unit feature vector $u_i\in\R^p$ (row $i$ of $U$). Every kernel
below is built from the linear kernel $K=UU\T$, $K_{ij}=\langle u_i,u_j\rangle$, which
on unit features is cosine similarity; because we compare selection methods rather
than kernels, we use this most basic choice throughout. With a ridge $\rho>0$, every category
kernel is positive definite (\cref{ass:spd}).

\paragraph{Shared kernel (Figure~\ref{fig:kernel-base}(a)).}
Although it is not part of our method, for comparison we begin with the traditional
construction, in which all categories share a single Gram matrix $K$, the
\emph{shared kernel}. Let $G_1,\ldots,G_M\subseteq[n]$ be nonempty categories. For
$S\subseteq[n]$, write $n_q(S)=|S\cap G_q|$. Memberships can overlap, so
$\sum_q n_q(S)$ may exceed $|S|$, while each selected item still occupies only one
of the $|S|$ places. With the shared kernel, every category sees the same
similarities and the categories play no role;
selecting within each category separately instead ignores that items can
serve several categories (\cref{supp:sec:independent-selection}).

\paragraph{Zero padding (Figure~\ref{fig:kernel-base}(b)).}
The simplest way to give each category a kernel over the whole dataset is to
keep the shared kernel on its members and set every other entry to zero:
$L_0^{(q)}=D_qKD_q$, where $D_q=\diag(\mathbf1\{i\in G_q\})$. As
\cref{prop:augmented-exact} shows, this changes nothing for a single category:
the padded kernel picks the same items as the shared kernel applied to the
category's members alone.
\begin{proposition}\label{prop:augmented-exact}
For $1\le k_q\le|G_q|$, the maximum of $\logdet((L_0^{(q)})_S)$ over all
size-$k_q$ subsets equals the maximum of $\logdet K_S$ over size-$k_q$ subsets of $G_q$.
If this maximum is finite, their maximizing sets coincide.
\end{proposition}
Zero padding, however, cannot handle several categories at once. A shared
subset almost always contains nonmembers of some category, and each nonmember
creates a zero row and column, so that category scores $-\infty$. Building on
its structure, we now introduce our softened category kernels.

\paragraph{Softened kernel (Figure~\ref{fig:kernel-soft}).}
Here, we replace the zero weights that zero padding gives nonmembers in $D_q$
with a small membership floor, and add a ridge:
\begin{equation}\label{eq:group-kernel}
L^{(q)}=\Omega_qK\Omega_q+\rho I_n,\qquad
\Omega_q=\diag(\omega_q(1),\ldots,\omega_q(n)),\quad
\omega_q(i)=\begin{cases}1&i\in G_q,\\ \LO&i\notin G_q,\end{cases}
\end{equation}
where $0\le\LO\le1$ and $\rho>0$. The ridge keeps every principal submatrix
positive definite. At the endpoint $\LO=0$, $\Omega_q=D_q$ and \eqref{eq:group-kernel}
reduces to zero padding plus a ridge,
$L^{(q)}=D_qKD_q+\rho I_n$: each nonmember
keeps only the diagonal entry $\rho$. With $\LO>0$
(Figure~\ref{fig:kernel-soft}(b)), nonmembers are down-weighted by $\LO$
instead of removed.
A multi-category item receives weight one in each of its categories. Every
candidate subset now has a finite score in every category.

The softened kernel balances category representation against feature
diversity, as its two endpoints show for $|S|=k$. With nonmembers removed ($\LO=0$) but
the ridge kept, writing $S_q=S\cap G_q$,
\begin{equation}\label{eq:block-main}
\logdet L^{(q)}_S
=\logdet(K_{S_q}+\rho I)+(k-n_q(S))\log\rho,
\end{equation}
that is, the diversity of the category's own selected members, plus a cost $\log\rho$
for each selected nonmember. With nonmembers faded ($\LO>0$) but no ridge
(Figure~\ref{fig:kernel-soft}(a)), and $K_S\succ0$,
factorization instead gives
\begin{equation}\label{eq:count-identity-main}
\logdet L^{(q)}_S
=2n_q(S)\log(1/\LO)+\logdet K_S+2k\log\LO,
\end{equation}
that is, a member count plus the diversity of the whole subset, shared by all
categories. Here the smallest of these scores across categories trades the count
floor $\min_q n_q(S)$ against shared feature volume; smaller $\LO$ increases the
count coefficient, and at $\LO=1$ every kernel is the same pooled kernel.  With both $\LO>0$ and
$\rho>0$, by \cref{lem:group-perturbation} the kernel stays close to
the category-specific regime \eqref{eq:block-main} when $\LO$ is small
relative to $\rho$, so a tiny floor does not make the score a pure
count.

All category kernels share $K$, $\LO$, and $\rho$, fixed before selection, so
their diversities are directly comparable (\cref{supp:rem:calibration}).\label{rem:calibration}
Logarithms are natural, and proofs are in \cref{sec:appendix-proofs}.

\section{A Soft Minimum, and Why It Is Needed}\label{sec:fair-objective}
We first define the soft minimum and how it scores one shared subset across all
categories. We then present two favorable properties: (1)~unlike the exact minimum,
it does not stall on ties between categories, and (2)~together with category
kernels it prefers items that serve several categories.

\paragraph{The soft-minimum objective.}
For one shared size-$k$ subset, let $d_q(S)=k^{-1}\logdet L^{(q)}_S$.
The maximin goal is $\max_{|S|=k}\min_qd_q(S)$. We optimize its smooth form
\begin{equation}\label{eq:fair-discrete}
\max_{|S|=k}F_\tau(S),\qquad
F_\tau(S)=-\tau\log\!\left[\frac1M\sum_{q=1}^M e^{-d_q(S)/\tau}\right].
\end{equation}
It satisfies $\min_qd_q(S)\le F_\tau(S)\le\min_qd_q(S)+\tau\log M$.
Smaller $\tau$ tracks the weakest category diversity more closely; with finite $\tau$,
stronger categories can offset a weak one only slightly, since $F_\tau$ never exceeds
the minimum by more than $\tau\log M$. Unlike a determinant of the averaged kernel \citep{chao2015largemargin}, it
retains each category's score. Unlike a fixed-weight sum of log-determinants \citep{ohsaka2020product},
it adapts emphasis toward categories with weaker diversity.

Equivalently, $F_\tau$ is the logarithm of a negative-order power mean of
$h_q(S)=\det(L^{(q)}_S)^{1/k}$ with exponent $\kappa=-1/\tau<0$. It emphasizes
weak category diversities rather than equality for its own sake, and unlike
the exact minimum it also responds to the other categories, which is why the
experiments report more than coverage alone.

\paragraph{(1) More favorable than the exact minimum.} The obvious way to lift the weakest
category is to maximize the exact minimum, the score of the single worst category.
However, the exact minimum creates ties that stall local search. With zero ridge and $K_S\succ0$, \eqref{eq:count-identity-main}
gives every category the same diversity term $\logdet K_S$, so categories differ
only through their integer member counts.
\begin{theorem}[Exact ties]\label{thm:ties}
Let $\rho=0$, $0<\LO<1$, and $K_S\succ0$. Then
$k\,d_q(S)=\logdet K_S-2\bigl(k-n_q(S)\bigr)\log\frac1\LO$, so $n_q(S)=n_r(S)$
implies $d_q(S)=d_r(S)$. Let $C^\star(S)$ be the categories attaining
$\min_qn_q(S)$. Adding an item that does not belong to every category in
$C^\star(S)$ leaves $\min_qn_q$ unchanged, so at the common next-set size
these candidates differ only through $\logdet K$. In contrast, for $\tau>0$
raising the count of any category in $C^\star(S)$ raises the soft minimum of
the count vector.
\end{theorem}
Under the exact minimum, whenever two or more categories tie at the lowest
count and no candidate belongs to all of them, the non-covering additions
have the same minimum-count contribution at their common resulting size.
A swap cannot raise the minimum count in this situation, but can lower it;
only among count-preserving swaps does diversity alone determine the ranking. With $M$ categories and a small subset, such ties are the
normal case. The soft minimum gives every tied category a positive weight, so
serving any of them is rewarded. On 50 COCO pools with eight categories, greedy
on $F_\tau$ ($\tau=0.05$ or $0.1$) scores higher on the \emph{exact} minimum
than greedy run on the exact minimum itself, in every pool
(\cref{supp:sec:strict}). With a small ridge the ties hold
approximately: at $\rho=10^{-10}$ and $\LO=10^{-3}$, COCO categories with equal
counts agree to six decimals.

\paragraph{(2) Shared items are preferred.}
The second property concerns our kernel design: category kernels (Figure~\ref{fig:kernel-soft}) favor, at equal
diversity gain, items shared by several categories, whereas the shared kernel
(Figure~\ref{fig:kernel-base}(a)) does not. This is desirable: with only $k$ places, each selected item should
serve as many categories as possible. For a current subset $S$ and a candidate $i\notin S$,
let $s_i>0$ be its \emph{diversity gain}, the Schur complement of $i$ against
the selected items (\cref{sec:appendix-shared}), assumed equal for every
category, and let $C_i$ be its set of categories.
\begin{proposition}[Shared items are preferred]\label{prop:shared}
Let $F_\tau$ be the soft-min objective \eqref{eq:fair-discrete}, and let $i,j\notin S$ have $s_i=s_j=s$ and $C_j\subsetneq C_i$. With the shared kernel $K+\rho I$ for every category
(Figure~\ref{fig:kernel-base}(a)), $F_\tau(S\cup\{i\})=F_\tau(S\cup\{j\})$.
With zero padding plus a ridge, or with the softened kernel without a ridge
(Figure~\ref{fig:kernel-soft}(a)),
\[
F_\tau(S\cup\{i\})-F_\tau(S\cup\{j\})
=\frac{\gamma}{|S|+1}\sum_{q\in C_i\setminus C_j}\bar w_q>0,
\]
where $\bar w_q\in(0,1]$ are soft-min weights at an intermediate point and
$\gamma=\log(1+s/\rho)$ or $\gamma=2\log(1/\LO)$, respectively.
\end{proposition}
In words, among equally diverse candidates, the objective prefers the item that
serves more categories, and the bonus is largest when those extra categories are
the weakest, because their soft-min weights are largest. With both a floor and a
ridge, the same preference holds approximately when $\LO$ is small relative to
$\rho$ (\cref{lem:group-perturbation}).

\section{Greedy Selection and Its Limits}\label{sec:greedy}
\paragraph{Algorithm.} With $\rho=0$ and $K_T\succ0$, the identity
\eqref{eq:count-identity-main} factorizes every category log-determinant into one
shared term and a count term:
$\logdet L^{(q)}_T=\logdet K_T+\sum_{i\in T}2\log\omega_q(i)$. For a size-$k$
set $S$ with $\lambda_{\min}(K_S)\ge\mu>0$, adding a category ridge changes
its normalized category scores by at most
$\log(1+\rho/(\LO^2\mu))$; thus $\rho\ll\LO^2$ alone is not a uniform
small-error condition. We therefore maximize
\[
\Phi_\tau(T)=\logdet(K_T+\epsilon I)+\SoftMin_{\tau k}\Bigl(\textstyle\sum_{i\in T}2\log\omega_q(i)\Bigr)_{q\in[M]},
\]
which ranks size-$k$ subsets like $F_\tau$ when $\rho=\epsilon=0$, greedily
(\cref{alg:greedy}); with positive ridges it is a surrogate, and nearly tied
rankings need not be preserved. Here
$\SoftMin_t(x)=-t\log\frac1M\sum_qe^{-x_q/t}$ and $\epsilon$ is a small
numerical ridge. Adding item $i$ to $T$ changes the first term by $\log s_i^{(\epsilon)}(T)$,
where $s_i^{(\epsilon)}(T)=K_{ii}+\epsilon-K_{iT}(K_T+\epsilon I)^{-1}K_{Ti}$ is
the Schur complement; write $s_i=s_i^{(0)}$ for the zero-ridge case. A single
Cholesky factorization of $K_T+\epsilon I$, shared by all categories, updates every
$s_i^{(\epsilon)}(T)$ in $O(n(p+t))$ per step, and the count term costs $O(nM)$. Greedy
therefore costs $O(nk(p+k+M))$ and never forms an $n\times n$ or category-specific
kernel. Optional one-item swaps against the pool refine the result.

\begin{algorithm}[t]
\caption{Greedy maximization of the soft minimum}\label{alg:greedy}
\begin{algorithmic}[1]
\Require unit features $u_i\in\R^p$; memberships $\omega_q(i)\in\{1,\LO\}$; size $k$; temperature $\tau$; numerical ridge $\epsilon\ge0$
\State $T\leftarrow\emptyset$;\quad $s_i^{(\epsilon)}\leftarrow\|u_i\|^2+\epsilon$ and $x_i\leftarrow\bigl(2\log\omega_q(i)\bigr)_q$ for all $i$;\quad $c\leftarrow0\in\R^M$
\For{$t=1,\dots,k$}
  \State $j\leftarrow\arg\max_{i\notin T}\ \log s_i^{(\epsilon)}+\SoftMin_{\tau k}(c+x_i)$
  \State $T\leftarrow T\cup\{j\}$;\quad $c\leftarrow c+x_j$
  \State update all $s_i^{(\epsilon)}$ by one Cholesky step for $K_T+\epsilon I$ with the new column $Uu_j$
\EndFor
\State \Return $T$ \Comment{optionally refined by one-item swaps}
\end{algorithmic}
\end{algorithm}

\paragraph{When greedy succeeds.} Under the condition below, on partitioned categories greedy is
coverage-optimal and its diversity loss is bounded by the feature geometry
alone.
\begin{proposition}[Greedy on partitions]\label{prop:greedy-partition}
Let the categories partition the items, each with at least $\lceil k/M\rceil$
members, let features have unit norm, $\rho=\epsilon=0$, and let $\mu>0$ be
the smallest eigenvalue of $K_A$ over all $1\le|A|\le k$. Put
$\beta=1-\LO^{2/(\tau k)}$ and $\Delta=\tau k\log(1+\beta^2/M)$. If
$e^{-\Delta}<\mu$, \cref{alg:greedy} returns a subset $T$ whose smallest
category count is $\lfloor k/M\rfloor$, the largest possible, and
$\min_qd_q(T)\ge\max_{|S|=k}\min_qd_q(S)-\log\frac1\mu$. With
$\tau=\log(1/\LO)/k$ the condition holds for every small enough $\LO$.
\end{proposition}
The loss $\log\frac1\mu$ does not depend on $\LO$. The soft minimum matters
here: under the exact minimum, tied categories stall greedy
(\cref{thm:ties}).

\paragraph{When greedy fails.} Greedy is the practical selector in our
experiments, but it has a characterizable failure mode. Call item $i$
\emph{covering} for $T$ if it belongs to every category with the lowest count.
\begin{lemma}[Greedy's choice rule]\label{lem:greedy-rule}
Under the exact minimum ($\tau=0$) and $\rho=\epsilon=0$, greedy picks the item
maximizing $\log s_i(T)+2\log\frac1\LO\cdot\mathbf 1\{i\text{ is covering}\}$. It
prefers a covering item $i$ to a non-covering item $j$ whenever
$s_i(T)>\LO^2\,s_j(T)$.
\end{lemma}
A covering item thus receives a fixed bonus $2\log\frac1\LO$ however redundant it
is. Near-duplicates of items that belong to several categories keep winning,
and the resulting subset can be stable under one-item swaps, because removing
one such item lowers several category counts at once.
\begin{proposition}[Greedy can lose $2(\tfrac1M-\tfrac1k)\log\tfrac1\LO$ per item]\label{prop:greedy-gap}
For fixed $M\ge2$ and $k=Mt>M$, there is a family of instances with
$\LO\to0$ on which greedy with $\tau=\rho=\epsilon=0$, followed by improving
one-item swaps, returns a subset whose $\min_qd_q$ is lower than that of
another subset by
$2\bigl(\frac1M-\frac1k\bigr)\log\frac1\LO-\frac{\log k}k-o(1)$.
Here $o(1)$ is taken as $\LO\to0$, with the auxiliary hub parameter
$\nu(\LO)\to0$ in the proof.
\end{proposition}
The instance has $k$ or more near-duplicate items, each belonging to every
category, and orthonormal single-category items (\cref{sec:appendix-greedy}). The loss grows
without bound as $\LO\to0$, so exact-minimum greedy has no constant-gap guarantee.
For partitions, \cref{prop:greedy-partition} excludes such a loss only under
its margin condition; it does not establish that overlap is necessary. In this
regime the Stiefel relaxation of \cref{supp:sec:contribution} can help: it
optimizes all $k$ columns at once, and its rounded solutions, refined by swaps, escaped the trap in the near-duplicate
instances we tested (\cref{supp:sec:solver-study}). We have no proof that it
always does. Keeping the better of the greedy and rounded subsets is never
worse than greedy (\cref{prop:safeguard}), but one \SCF\ sweep costs as
much as a whole greedy run when $M\ge2$ (\cref{prop:cost}).

\section{When the Relaxation Earns Its Cost}\label{sec:solver-comparison}
The exact-minimum construction in \cref{prop:greedy-gap} identifies a
regime to test with the soft minimum. Shared labels reward a near-duplicate
item for covering several weak categories. Greedy can run out of budget
before it can diversify those categories, and a swap can sacrifice coverage
in several categories at once. The relaxation can change all $k$ directions
jointly. That leaves two obstacles. The nonlinear iteration must make
progress, and rounding must preserve enough of that progress to improve
the discrete subset.

\paragraph{Safeguarded SCF.}
Start from a declared greedy baseline $S_{\mathrm g}$, with the prescribed
exchanges if used. Independently solve the relaxation, round it and refine
the resulting subset $S_{\mathrm r}$. Return $S_{\mathrm r}$ only if its
exact score $F_\tau(S_{\mathrm r})$ exceeds $F_\tau(S_{\mathrm g})$.
Otherwise, return $S_{\mathrm g}$. This rule guarantees
$F_\tau(S_{\mathrm{out}})\ge F_\tau(S_{\mathrm g})$ for every pool
(\cref{prop:safeguard}). The guarantee is relative to the declared
baseline and the exact discrete objective. It costs both selection
stages and two score evaluations. A selection that is already globally
optimal can only tie, and probe accuracy has a separate objective.

\paragraph{Shift and acceleration.}
\Cref{prop:gap} gives a gap of at least $\beta=1/k+\sigma$ for the
\emph{frozen} lifted matrix. At a fixed frame, increasing $\sigma$ increases
this gap and decreases the rotation toward the top eigenspace. An SCF
objective comparison uses the nonlinear objective evaluated at two
different frames. For the soft-min log-determinant, we prove that a
sufficiently large $\beta$ makes the frozen trace gain a lower bound for
the true objective gain, as required by Li's NEPv Ansatz
(\cref{prop:shift-ansatz}). The threshold depends on objective curvature.
We test LOCG \citep{li2024nepv,wang2026nepvadmm} with five SCF sweeps in
$\operatorname{span}[V_t,A_\perp(V_t),V_{t-1}]$. Each sweep refreshes the
category weights using exact projected kernels. The baseline uses the
original $2k$ compressed eigensolve with matched shift and damping.
On the COCO shift sweep, the unshifted run had a minimum gap of $0.010$
and 48 objective decreases in 100 updates. At $k\sigma=2$ the minimum
gap was $0.030$ and every step was nondecreasing at tolerance $10^{-9}$.
At $k\sigma=80$ the gap reached at least $0.810$, but the final
objective was lower than at $k\sigma=2$ ($-8.261$ vs. $-8.030$).
\Cref{supp:sec:locg} gives the full protocol and diagnostics.

\paragraph{Dense overlap and near-duplicates.}\label{sec:near-dup}
In the earlier VisDrone study, SCF with leverage rounding and swaps beat
the best of 12 greedy temperatures and 200 random starts with swaps on
all 15 tested pools at $k=20$ and both reported floors (\cref{tab:visdrone}).
We rerun the comparison against plain greedy and greedy with exchanges
on matched pools, then test neighboring budgets with archived multistart
scores as a stronger comparator. The RTX 5080 study separates solver
iterations and timing from the objective on the returned subset.
\begin{table}[!htbp]
\centering
\caption{\textbf{Matched RTX 5080 solver comparison.}
VisDrone uses three pools per floor ($k=20$). COCO uses one pool ($k=100$).
The relaxed and discrete objectives are means. Residual and elapsed times
are medians. Selection time includes rounding and swaps. It starts
with loaded features and initial frames, before any probe fitting.
SCF and LOCG use the GPU for the relaxation. Greedy and swaps use
the CPU. mAP is in percent.}
\label{tab:locg-main}
\footnotesize
\setlength{\tabcolsep}{5pt}
\begin{tabular}{llrrrrr}
\toprule
Benchmark & Method & Residual & Relaxed $f_\tau$ & Discrete $F_\tau$ & mAP & Select (s)\\
\midrule
VisDrone, $\LO=0.03$ & Greedy & -- & -- & $-5.802$ & -- & 0.01 \\
 & Greedy + swaps & -- & -- & $-5.620$ & -- & 0.05 \\
 & Adaptive SCF & $3.1\!\times\!10^{-3}$ & $-1.373$ & $-5.513$ & -- & 5.59 \\
 & SCF + LOCG & $5.2\!\times\!10^{-4}$ & $-1.370$ & $-5.512$ & -- & 22.85 \\
\midrule
VisDrone, $\LO=0.05$ & Greedy & -- & -- & $-5.206$ & -- & 0.01 \\
 & Greedy + swaps & -- & -- & $-5.058$ & -- & 0.05 \\
 & Adaptive SCF & $2.7\!\times\!10^{-3}$ & $-1.175$ & $-5.001$ & -- & 5.67 \\
 & SCF + LOCG & $3.3\!\times\!10^{-4}$ & $-1.173$ & $-5.007$ & -- & 23.01 \\
\midrule
COCO & Greedy & -- & -- & $-13.973$ & 34.94 & 0.63 \\
 & Adaptive SCF & $4.3\!\times\!10^{-4}$ & $-8.030$ & $-14.150$ & 32.79 & 47.73 \\
 & SCF + LOCG & $8.7\!\times\!10^{-5}$ & $-8.029$ & $-14.139$ & 32.46 & 236.18 \\
\bottomrule
\end{tabular}
\end{table}

At $k=20$ both relaxed solvers beat greedy plus swaps in all six paired
cases (\cref{tab:locg-main}) and also beat their archived multistart
comparators. The mean SCF gains over greedy plus swaps
are $0.107$ and $0.056$ at the two floors. LOCG reaches the terminal SCF
residual in $4.0$--$5.1$ seconds, compared with $5.5$--$5.6$ seconds for
SCF. At the common cap, LOCG has lower residual and higher relaxed
objective in all six cases. It takes about four times as long. After
rounding it wins one, ties two and loses three against SCF.

\paragraph{Budget boundary.}
The five additional pools average $5.4$--$5.5$ labels per frame and
have median nearest-neighbor squared sines of $0.0067$--$0.0070$.
At $k=20$, both relaxed solvers beat the stronger archived multistart
baseline on all five pools, adding mean score gains of $0.049$ for SCF
and $0.052$ for LOCG. Neither has a strict win against that comparator
at $k=15$. At $k=25$, SCF wins three pools and LOCG wins two, with
mean differences below $0.001$ in magnitude. Thus the sizeable
advantage is concentrated at the intermediate budget
(\cref{tab:locg-budgets}). Against freshly measured greedy plus swaps,
the reconstructed safeguard gives 19 strict gains and two ties across
the 21 VisDrone pairs with either relaxed solver. This reconstruction
uses exact saved scores. \Cref{tab:locg-main} reports measured component
times. VisDrone has no downstream probe in this comparison.

\paragraph{COCO contrast.}
On the one-eighth COCO pool, SCF takes $47.7$ seconds through rounding
and scores $-14.150$. LOCG lowers the residual by a factor of $4.9$
and raises the discrete objective to $-14.139$ in $236.2$ seconds,
with a slightly lower mAP. Greedy scores $-13.973$ and reaches $34.94\%$
mAP in $0.63$ seconds of selection time. The rare-category mAP values
for greedy, SCF and LOCG are $23.27\%$, $20.20\%$ and $18.21\%$.
The safeguard retains greedy in this split. The full-pool results in
\cref{supp:sec:solver-study} also favor greedy.

\section{Downstream Results}\label{sec:experiments-real}
We ask where our method trains a better downstream model than random
selection and than class-balanced sampling, and where it does not. Unless stated
otherwise, experiments share the protocol below; \cref{supp:sec:lt-details}
describes the data, settings and aggregated results.

\paragraph{Protocol.}
Every item is represented by its CLIP ViT-B/32 image embedding (OpenAI weights, OpenCLIP
\texttt{ViT-B-32-quickgelu}; \citealp{radford2021clip,ilharco2021openclip}), normalized to unit
length, and every selector uses the linear kernel of \cref{sec:model} on these features.
We compare seven selectors on the same pool and features:
uniform sampling; class-balanced sampling, which repeatedly adds a random unselected item from
the category with the fewest selected items; $k$-center (farthest-first traversal in cosine
distance from a random start); greedy MAP on the shared kernel $K$ (shared-kernel DPP); greedy
MAP on the averaged category kernel $\frac1M\sum_qL^{(q)}$; greedy maximization of the
fixed-weight average $\frac1M\sum_qd_q$; and ours, \cref{alg:greedy} on $F_\tau$. The selected
items train a logistic-regression probe on the frozen features with identical settings for all
selectors (300 full-batch AdamW epochs, learning rate $10^{-2}$, weight decay $10^{-4}$, inputs
scaled by 10). Each configuration uses five seeds, which draw the random selectors and initialize the probe;
we report mean, standard deviation and the number of seeds on which ours is higher. The
floor $\LO$, shared by the three category-kernel selectors, and temperature $\tau$
were chosen on one development setting.

\paragraph{Choosing the floor.}
With a negligible ridge, \eqref{eq:count-identity-main} shows that one extra member changes a
category diversity by $2\log(1/\LO)/k$. If this is smaller than the temperature, the soft
minimum averages over categories and stops protecting empty ones: with the COCO setting
$\LO=10^{-3}$, $\tau=0.1$ and $k=400$, the step is $0.035<\tau$, and on CIFAR-100-LT (below) our
subset left 11 of 100 classes empty. On that development setting (\cref{tab:app-dev}),
lowering $\tau$ below $0.03$ made coverage worse, because a near-hard minimum ties all empty classes, whereas
lowering $\LO$ restored it. We fix $\LO=10^{-30}$, $\tau=0.1$ and $\rho=10^{-64}\ll\LO^2$ for all
later runs, started only after these values were fixed.

\subsection{Long-tailed classification}\label{sec:lt}
We build CIFAR-10-LT and CIFAR-100-LT in the standard way \citep{krizhevsky2009cifar,cao2019ldam}:
with imbalance ratio $r$, class $i$ keeps the first $\lfloor n_{\max}r^{-i/(C-1)}\rfloor$ training
images, so CIFAR-100-LT with $r=100$ ranges from 500 to 5 images per class (10{,}847 images).
The test sets are the full, balanced CIFAR test sets, so test accuracy is balanced accuracy.
GTSRB \citep{stallkamp2012gtsrb} is used with its natural imbalance (43 classes; a fixed random
subset of 12{,}512 training and 4{,}000 test images); its training images come in tracks of
near-identical frames of one physical sign. ``Tail'' denotes the rarest 30\% of classes in the
pool. CIFAR-100-LT with $r=100$ is the development setting; CIFAR-10-LT ($r=100$) and GTSRB
were run with the settings frozen.

\begin{table}[t]
\centering
\caption{\textbf{Long-tailed classification: balanced test accuracy (\%)}, linear probe, mean{\scriptsize$\pm$}std over five seeds. $\LO$ and $\tau$ were chosen on CIFAR-100-LT; the other two use frozen settings.}
\label{tab:lt-acc}
\footnotesize
\setlength{\tabcolsep}{2.6pt}
\begin{tabular}{lcccccc}
\toprule
& \multicolumn{2}{c}{CIFAR-10-LT ($r{=}100$)} & \multicolumn{2}{c}{CIFAR-100-LT ($r{=}100$)} & \multicolumn{2}{c}{GTSRB}\\
\cmidrule(lr){2-3}\cmidrule(lr){4-5}\cmidrule(lr){6-7}
Selector & $k{=}200$ & $k{=}400$ & $k{=}200$ & $k{=}400$ & $k{=}200$ & $k{=}400$\\
\midrule
Uniform & 69.6{\scriptsize$\pm$5.7} & 71.6{\scriptsize$\pm$6.4} & 27.0{\scriptsize$\pm$0.9} & 33.7{\scriptsize$\pm$1.2} & 42.0{\scriptsize$\pm$2.0} & 50.3{\scriptsize$\pm$2.7}\\
Class-balanced & \textbf{90.6{\scriptsize$\pm$0.4}} & \textbf{91.3{\scriptsize$\pm$0.1}} & \textbf{43.7{\scriptsize$\pm$0.7}} & \textbf{54.2{\scriptsize$\pm$0.9}} & 49.8{\scriptsize$\pm$2.1} & 61.5{\scriptsize$\pm$1.5}\\
$k$-center & 73.0{\scriptsize$\pm$3.2} & 78.1{\scriptsize$\pm$1.0} & 28.4{\scriptsize$\pm$0.9} & 35.6{\scriptsize$\pm$1.0} & 39.1{\scriptsize$\pm$1.8} & 50.8{\scriptsize$\pm$1.4}\\
Shared-kernel DPP & 73.1{\scriptsize$\pm$0.1} & 83.1{\scriptsize$\pm$0.0} & 27.3{\scriptsize$\pm$0.0} & 33.6{\scriptsize$\pm$0.1} & 41.4{\scriptsize$\pm$0.1} & 49.5{\scriptsize$\pm$0.2}\\
Averaged kernel & 88.0{\scriptsize$\pm$0.1} & 89.8{\scriptsize$\pm$0.1} & 39.7{\scriptsize$\pm$0.1} & 50.8{\scriptsize$\pm$0.1} & \textbf{54.8{\scriptsize$\pm$0.3}} & \textbf{62.3{\scriptsize$\pm$0.2}}\\
Fixed weight & 70.0{\scriptsize$\pm$0.2} & 80.1{\scriptsize$\pm$0.2} & 27.0{\scriptsize$\pm$0.1} & 34.1{\scriptsize$\pm$0.2} & 42.1{\scriptsize$\pm$0.1} & 49.3{\scriptsize$\pm$0.5}\\
Ours (greedy) & 89.4{\scriptsize$\pm$0.1} & 90.5{\scriptsize$\pm$0.0} & 42.6{\scriptsize$\pm$0.1} & 53.4{\scriptsize$\pm$0.2} & 44.6{\scriptsize$\pm$0.1} & 56.3{\scriptsize$\pm$0.3}\\
\bottomrule
\end{tabular}
\end{table}

\Cref{tab:lt-acc} shows the main result. Against uniform sampling our subsets raise balanced
accuracy by 19.0--19.7 points on CIFAR-10-LT, 15.6--19.7 points on CIFAR-100-LT and 2.6--6.0
points on GTSRB, on at least four of five seeds in every case; tail-class accuracy rises from
2--35\% to 40--96\% (\cref{tab:lt-tail}). Category-blind diversity and fixed weights gain at most 11.6 points, because
they leave rare classes empty; with the COCO floor $\LO=10^{-3}$ our method gains only 5.0--5.3
points on CIFAR-100-LT (\cref{tab:app-dev}), so the gain needs the count regime.

\subsection{Rare classes and the class-balanced baseline}\label{sec:lt-cb}
Class-balanced sampling, which equalizes counts as our objective does in the count regime, is
the strongest of these baselines (\cref{supp:sec:extra-baselines} adds two more). \Cref{tab:lt-all} compares ours with both baselines in every setting,
marked as used to choose $\LO$ (develop.), run with frozen settings (confirm.), or added to test
one hypothesis registered before running them (pre-reg.): that ours is more accurate than
class-balanced sampling on tail classes. The hypothesis holds by our rule (higher mean and at least four of
five seeds) in 5 of these 8 cells; the mean is higher in 7, and in one cell (CIFAR-100-LT,
$r=50$, $k=200$) it is clearly lower. Overall accuracy is lower than class-balanced sampling by
0.7--1.6 points in 11 of the 12 CIFAR rows (0.2 points higher in the other) and by about 5
points on GTSRB, where the averaged-kernel selector is the most accurate of all. Ours is not better than class-balanced
sampling overall; it trades accuracy on frequent classes for accuracy on the rarest. 

\paragraph{Why class-balanced sampling is so close.}
If nonmembers are removed ($\LO=0$, zero padding) and $\rho\to0$, \eqref{eq:block-main} gives
$d_q(S)=\frac{n_q(S)}{k}\log\frac1\rho+\frac1k\logdet K_{S_q}+\log\rho+o(1)$: maximizing the weakest
category first raises the smallest count to that of class-balanced sampling, and then spreads out
the weakest category's own members (\cref{prop:zero-cb}). Class-balanced sampling thus reaches the same smallest count with
random instead of diverse picks, and our floor also puts counts first, so \cref{tab:lt-all}
measures what diversity adds at the same smallest count: it helps the rarest classes, and with few
items per class, diversity within each class helps further (\cref{supp:sec:extra-baselines}).

\paragraph{Full fine-tuning and near-duplicate pools.}
Neither changes this picture (\cref{tab:ft-dup}). Fine-tuning the whole CLIP image encoder on
the same subsets is within 0.3 points of class-balanced sampling or above it, and 12--23 points above
uniform. When every pool image also appears as two lightly augmented near-copies (cosine 0.97),
ours ties class-balanced sampling overall (52.1\% vs 51.7\% at $k=400$) and is more accurate on
tail classes (56.6\% vs 50.8\%, five of five seeds).

\subsection{Limits of the objective}\label{sec:lvis}\label{sec:coco}
\emph{Cluttered multi-label images.} On COCO 2017 \citep{lin2014coco} (80 categories,
$k\in\{100,200,400\}$, multi-label probe; \cref{supp:sec:coco-setup}), ours gives the highest
objective and largest smallest category, but uniform and class-balanced sampling train better
probes (\cref{tab:coco-main}). With $\LO=10^{-30}$ the smallest category reaches 5, 11 and 22
images (class-balanced: 2, 4.8, 10), yet mAP stays lower (\cref{tab:app-coco30}): the objective
favors images with about six labels (\cref{prop:shared}), cluttered for a linear probe.

\emph{Many more categories than picks.} On LVIS \citep{gupta2019lvis} (1{,}203 categories, $k=400$,
registered protocol with $\LO=10^{-3}$), ours beats uniform sampling on rare categories (AP$_r$ 5.9 vs 5.5) but loses
to class-balanced sampling everywhere, covering 14 rare categories against 164 (\cref{tab:lvis}):
with about 1{,}000 empty categories tied, covering one moves the soft minimum by only
about $2\log(1/\LO)/(1000k)\approx3\times10^{-5}$. Our method suits settings with fewer categories than picks.

\section{Discussion and Future Work}\label{sec:conclusion}
Shared-label redundancy is a regime in which the relaxed solver can earn
its cost. On VisDrone, using the relaxed frame as a starting point for
exchanges improves the subset score at the intermediate budget. The
comparison across budgets shows that shared labels and redundant features
are insufficient to predict a win.
The exact-minimum construction explains one way greedy can fail, while
the soft-min experiments measure its practical relevance.

Continuous and discrete performance should be judged separately.
A sufficient shift guarantees objective ascent for the relaxed update,
and LOCG can reach a given full-space residual sooner. COCO shows how a
better relaxed solution can still yield a weaker subset than greedy.
When subset score is the priority, evaluating both discrete candidates
and keeping the stronger one preserves greedy's value at the cost of
both runs. When throughput or the downstream probe is the priority,
greedy has the advantage on the measured COCO split. Its downstream gains over uniform
selection are clearest in the long-tailed count regime, while class-balanced
sampling remains competitive and the gains do not transfer to cluttered
multi-label COCO or to LVIS with many more categories than picks.
Future work should improve how the relaxation is rounded and test when
subset-score gains transfer to training on redundant video.

\clearpage
\section*{Acknowledgments}
This work received partial support from the Research Grants Council of Hong Kong
through General Research Fund grant 12300322.

\section*{Ethics Statement}
Fairness here refers to category-kernel scores, not demographic parity or
protected-group representation. The experiments use public benchmarks (CIFAR-10,
CIFAR-100, GTSRB, COCO, LVIS and VisDrone); public availability does not remove dataset licensing, privacy,
or downstream-use obligations, which must be checked for any release.

\section*{Reproducibility Statement}
\Cref{sec:model,sec:fair-objective,sec:greedy} specify the kernels, objective
and greedy selector, \cref{sec:contribution} the relaxed solver and rounding
rule, and \cref{sec:appendix-proofs,sec:contribution} provide the proofs.
\Cref{supp:sec:lt-details} lists the data sources, long-tailed
construction, features, selectors, training and evaluation settings, the order
in which settings were fixed, and complete per-dataset results;
\cref{supp:sec:solver-study} covers COCO and drone video, and
\cref{supp:sec:coco-setup} the solver parameters. We will release the code, data preparation and
feature-extraction scripts, split construction, selectors, selected subsets,
and per-seed outputs for the downstream experiments.
\Cref{supp:sec:locg} specifies the GPU solver comparison, for which we will
also release the scripts, input hashes, paired traces and selected
row indices.

\bibliography{references}
\bibliographystyle{iclr2027_conference}
\clearpage
\appendix
\section{Technical Appendix}\label{sec:appendix-proofs}
This appendix proves the results of \cref{sec:model,sec:fair-objective,sec:greedy}: the category-kernel
identities (\cref{sec:appendix-group-kernels}), shared items (\cref{sec:appendix-shared}), the relaxation
(\cref{sec:appendix-deferred,sec:appendix-gap}) and greedy (\cref{sec:appendix-greedy}); \cref{sec:contribution} states and proves the
solver results. Throughout,
$V\in\Stiefel(n,k)$,
\[
 L^{(q)}_V=V\T L^{(q)}V\succ0,\qquad
 d_q(V)=\tfrac1k\logdet L^{(q)}_V,\qquad
 P=VV\T,
\]
and $\langle X,Z\rangle=\tr(X\T Z)$ is the Frobenius inner product.
\Cref{ass:spd} guarantees $L^{(q)}_V\succ0$, because
$y\T L^{(q)}_{V}y=(Vy)\T L^{(q)}(Vy)>0$ for $y\ne0$ (and $Vy\ne0$ since $V\T V=I_k$).

\subsection{Category-kernel identities}\label{sec:appendix-group-kernels}
Categories $G_q$ may overlap, $n_q(S)=|S\cap G_q|$, $|S|=k$, and $K$ is the
shared kernel of unit feature vectors. All identities hold one category at
a time.

\begin{proof}[Proof of \cref{prop:augmented-exact}]
If $S\not\subseteq G_q$, some selected item indexes a zero row and column of
$(L_0^{(q)})_S$, so its determinant is $0$ and the log-determinant is
$-\infty$. If $S\subseteq G_q$, then $(L_0^{(q)})_S=K_S$. Hence every set
with a finite value lies in $G_q$, where the two objectives coincide.
\end{proof}

\begin{lemma}[Zero floor]\label{lem:block-ridge}
If $\LO=0$ and $\rho>0$, write $S_q=S\cap G_q$. After permuting $S$,
\begin{equation}\label{eq:block-ridge}
 L^{(q)}_S=\begin{pmatrix}K_{S_q}+\rho I&0\\0&\rho I_{k-n_q(S)}\end{pmatrix},
 \qquad
 \logdet L^{(q)}_S=\logdet(K_{S_q}+\rho I)+(k-n_q(S))\log\rho .
\end{equation}
Replacing a selected nonmember by an unselected member $i$ raises this
score by $\log(1+s/\rho)>0$, where
$s=K_{ii}-K_{iS_q}(K_{S_q}+\rho I)^{-1}K_{S_qi}>0$.
\end{lemma}
\begin{proof}
With $\LO=0$, $\Omega_S=\diag(I_{n_q},0)$, so $\Omega_SK_S\Omega_S=\diag(K_{S_q},0)$; adding
$\rho I$ and taking the determinant of the block-diagonal matrix gives
\eqref{eq:block-ridge}. Adding member $i$ multiplies the in-group determinant
by its Schur complement $\rho+s$, and removing a nonmember divides out one
factor $\rho$. Finally $s=u_i\T\rho(U\T U+\rho I)^{-1}u_i>0$, where the rows
of $U$ are the feature vectors of $S_q$.
\end{proof}

\begin{lemma}[Zero ridge; Figure~\ref{fig:kernel-soft}(a)]\label{lem:quality-factor}
If $0<\LO\le1$, $\rho=0$, and $K_S\succ0$, then
\begin{equation}\label{eq:quality-factor}
 \logdet L^{(q)}_S=2n_q(S)\log(1/\LO)+\logdet K_S+2k\log\LO .
\end{equation}
\end{lemma}
\begin{proof}
With $\rho=0$, $L^{(q)}_S=\Omega_SK_S\Omega_S$ for $\Omega_S=\diag(\omega_q(i):i\in S)$, so
$\det L^{(q)}_S=\det(\Omega_S)^2\det K_S=\LO^{2(k-n_q(S))}\det K_S$, because
exactly $k-n_q(S)$ diagonal entries of $\Omega_S$ equal $\LO$. Taking logarithms
with $2(k-n_q)\log\LO=2n_q\log(1/\LO)+2k\log\LO$ gives the claim.
\end{proof}
Since $\logdet K_S$ and $2k\log\LO$ do not depend on $q$, the minimum over
categories of \eqref{eq:quality-factor} is
$2\log(1/\LO)\min_qn_q(S)+\logdet K_S+2k\log\LO$: a count floor plus a shared
volume. For a partition, $\sum_qn_q(S)=k$ gives
$\min_qn_q(S)\le\lfloor k/M\rfloor$; with overlap, $\sum_qn_q(S)$ can exceed
$k$ and this ceiling no longer applies.

\begin{lemma}[Ridge-dominant perturbation]\label{lem:group-perturbation}
Let $0\le\LO\le1$, $\rho>0$, and let $A_0$ be the block matrix in
\eqref{eq:block-ridge}. If $\LO\sqrt{n_q(k-n_q)}+\LO^2(k-n_q)\le\zeta\rho$ with
$\zeta<1$, then
\begin{equation}\label{eq:group-perturbation}
 \bigl|\logdet L^{(q)}_S-\logdet A_0\bigr|\le-k\log(1-\zeta).
\end{equation}
\end{lemma}
\begin{proof}
Write $L^{(q)}_S=A_0+\Delta$ with
$\Delta=\bigl(\begin{smallmatrix}0&\LO C\\\LO C\T&\LO^2K_{S_{-q}}\end{smallmatrix}\bigr)$,
where $C$ holds inner products between members and nonmembers in $S$. Since
all entries have magnitude at most one,
$\|C\|_2\le\sqrt{n_q(k-n_q)}$ and $\|K_{S_{-q}}\|_2\le k-n_q$, so
$\|A_0^{-1}\Delta\|_2\le\|\Delta\|_2/\rho\le\zeta$. The eigenvalues of
$A_0^{-1}\Delta$ are real (it is similar to $A_0^{-1/2}\Delta A_0^{-1/2}$)
and lie in $[-\zeta,\zeta]$, so
$|\logdet(I+A_0^{-1}\Delta)|\le-k\log(1-\zeta)$.
\end{proof}
When $\LO\ll\rho/k$, the zero-floor expansion \eqref{eq:block-ridge},
not the zero-ridge identity \eqref{eq:quality-factor}, describes the score.

\subsection{Shared items are preferred}\label{sec:appendix-shared}
Throughout, category diversities of the candidate sets $S\cup\{i\}$ are
normalized by their common size $|S|+1$. Write $D_q(T)=\logdet L^{(q)}_T$ and
$\Delta_q(i)=D_q(S\cup\{i\})-D_q(S)$ for the change caused by adding $i$.

\paragraph{The gain of one item in each kernel.} Adding a row and column to a
positive-definite matrix multiplies its determinant by the Schur complement of
the new entry.
\begin{itemize}[leftmargin=*]
\item \emph{Shared kernel, $L^{(q)}=K+\rho I$.} Every category sees the same
matrix, so $\Delta_q(i)=\log(\rho+s_i)$ with
$s_i=K_{ii}-K_{iS}(K_S+\rho I)^{-1}K_{Si}$, for every $q$.
\item \emph{Zero padding plus ridge.}
By \cref{lem:block-ridge}, $L^{(q)}_S$ is block diagonal. A member of $q$
joins the member block and contributes its Schur complement
$\rho+s_i$, with $s_i=K_{ii}-K_{iS_q}(K_{S_q}+\rho I)^{-1}K_{S_qi}$; a
nonmember adds a diagonal entry $\rho$. Hence $\Delta_q(i)=\log(\rho+s_i)$ if
$q\in C_i$ and $\log\rho$ otherwise. Here $s_i$ is measured against the members
$S_q$ of each category; the proposition assumes it takes the same value in
every category.
\item \emph{Softened kernel without ridge (Figure~\ref{fig:kernel-soft}(a)).} By \cref{lem:quality-factor}, $D_q(T)=2\sum_{l\in
T}\log\omega_q(l)+\logdet K_T$, so
$\Delta_q(i)=2\log\omega_q(i)+\log s_i$ with
$s_i=K_{ii}-K_{iS}K_S^{-1}K_{Si}$, which is the same for every category.
Hence $\Delta_q(i)=\log s_i$ if $q\in C_i$ and $\log s_i-2\log(1/\LO)$
otherwise.
\end{itemize}
In each case $\Delta_q(i)=a$ for member categories and $b$ for the others,
with $a-b=0$, $\log(1+s_i/\rho)$, and $2\log(1/\LO)$, respectively.

\begin{proof}[Proof of \cref{prop:shared}]
With $s_i=s_j$, the gains of $i$ and $j$ agree on every category except those
in $C_i\setminus C_j$, where $\Delta_q(i)-\Delta_q(j)=a-b=\gamma$. Let
$x_q=D_q(S\cup\{j\})/(|S|+1)$ and $y_q=D_q(S\cup\{i\})/(|S|+1)$, so
$y-x=\frac{\gamma}{|S|+1}\mathbf 1_{C_i\setminus C_j}$. The soft minimum
$\phi(x)=-\tau\log\frac1M\sum_qe^{-x_q/\tau}$ is smooth with gradient equal
to the soft-min weights, which lie in $(0,1]$. By the mean value theorem,
$\phi(y)-\phi(x)=\sum_q\bar w_q(y_q-x_q)$ for the weights $\bar w$ at a point
between $x$ and $y$, which gives the stated difference. For the shared kernel
$\gamma=0$, so the two values coincide.
\end{proof}

\subsection{From subsets to orthonormal frames}\label{sec:appendix-deferred}
\begin{proof}[Proof of \cref{lem:relaxation}]
A coordinate frame $V_S$ has distinct standard basis vectors as columns, so
$V_S\T V_S=I_k$ and $(V_S\T L^{(q)}V_S)_{j\ell}=L^{(q)}_{i_ji_\ell}$. Hence
$L^{(q)}_{V_S}=L^{(q)}_S$ and $f_\tau(V_S)=F_\tau(S)$, and maximizing over all
frames can only increase the value. For orthogonal $Q$,
$L^{(q)}_{VQ}=Q\T L^{(q)}_{V}Q$ has the same determinant, so $f_\tau(VQ)=f_\tau(V)$.
\end{proof}

\begin{proposition}[One kernel]\label{prop:spectral-relaxation}
If $L\succeq0$ has eigenvalues $\lambda_1\ge\cdots\ge\lambda_n$ and
$\lambda_k>0$, then
$\max_{V\in\Stiefel(n,k)}k^{-1}\logdet(V\T LV)=k^{-1}\sum_{i\le k}\log\lambda_i$,
attained by any leading $k$-dimensional eigenspace.
\end{proposition}
\begin{proof}
The multiplicative Rayleigh--Ritz principle gives
$\det(V\T LV)\le\prod_{i\le k}\lambda_i$, with equality on a leading
eigenspace. Take logarithms and divide by $k$.
\end{proof}

\subsection{The relaxation is loose for small floors}\label{sec:appendix-gap}
In a coordinate frame each column is one item, which counts as a member only
for its own categories. A general orthonormal frame can instead superpose items
from several categories in one column, so that every category sees a member in
it. The relaxation can therefore avoid most of the nonmember penalty, and its
optimum separates from every discrete subset as the floor shrinks.

\begin{proposition}[Integrality gap]\label{prop:relax-gap}
Let the categories partition $[n]$, let every $u_i$ have unit norm, and let
$\rho=0$. Then:
\begin{enumerate}[leftmargin=*]
\item for every size-$k$ subset $S$,
$\displaystyle\min_q d_q(S)\le-2\Bigl(1-\tfrac{\lfloor k/M\rfloor}{k}\Bigr)\log(1/\LO)$;
\item if each category has $k$ members $T_q\subseteq G_q$ with $K_{T_q}\succ0$,
$\displaystyle\liminf_{\LO\to0}\ \max_{V\in\Stiefel(n,k)}\min_q d_q(V)\ge-\log M
+\min_q\tfrac1k\logdet K_{T_q}$.
\end{enumerate}
\end{proposition}
\begin{proof}
(1) By \cref{lem:quality-factor},
$d_q(S)=\frac1k\logdet K_S-\frac{2(k-n_q(S))}{k}\log(1/\LO)$. For a partition,
some category has $n_q(S)\le\lfloor k/M\rfloor$, and Hadamard's inequality gives
$\logdet K_S\le\sum_{i\in S}\log K_{ii}=0$. (If $K_S$ is singular the left side
is $-\infty$.)

(2) Write $T_q=\{i_{q,1},\ldots,i_{q,k}\}$ and let column $j$ of $V$ be
$v_j=M^{-1/2}\sum_qe_{i_{q,j}}$. The columns have disjoint supports, so $V$ is
orthonormal. Since the categories partition $[n]$,
$\Omega_qv_j=M^{-1/2}\bigl(e_{i_{q,j}}+\LO\sum_{q'\ne q}e_{i_{q',j}}\bigr)
\to M^{-1/2}e_{i_{q,j}}$ as $\LO\to0$, hence
$L^{(q)}_V=V\T\Omega_qK\Omega_qV\to\frac1MK_{T_q}\succ0$. By continuity of
$\logdet$ on positive-definite matrices,
$d_q(V)\to-\log M+\frac1k\logdet K_{T_q}$ for every $q$, and the maximum over
$V$ is at least this value.
\end{proof}
For $M\ge2$, the bound in (1) decreases proportionally to $\log(1/\LO)$, while
the limiting lower bound in (2) does not depend on $\LO$, so the gap grows
without bound as $\LO\to0$. Because
$\min_qd_q\le F_\tau\le\min_qd_q+\tau\log M$, for fixed $\tau$ the same holds for $F_\tau$ up
to $\tau\log M$. A relaxed solution can thus reach a high value through
superposed columns that no subset can imitate, which makes it a weak guide for
rounding when $\LO$ is small.

\paragraph{Numerical check.} With $n=600$ unit features in $\R^{40}$, $M=5$
equal categories and $k=12$, the best of greedy, greedy with swaps, and 200
random subsets stays below the bound in (1) at every floor tested (for example
$-11.53$ against $-11.51$ at $\LO=10^{-3}$, and $-23.04$ against $-23.03$ at
$\LO=10^{-6}$), so the bound is nearly tight. The superposed frame of (2)
reaches $-1.82$ at every floor from $10^{-2}$ down, matching the predicted limit.

\subsection{Exact ties and the limits of greedy}\label{sec:appendix-greedy}
Throughout this subsection $\rho=0$, features have unit norm, $L'=\log\frac1\LO$,
and for a set $T$ we write
$\Psi(T)=\logdet K_T+2L'\min_qn_q(T)$. By \cref{lem:quality-factor},
$|T|\min_qd_q(T)=\Psi(T)-2|T|L'$, so for sets of equal size the exact minimum and
$\Psi$ rank subsets identically.

\begin{proof}[Proof of \cref{thm:ties}]
\Cref{lem:quality-factor} gives
$k\,d_q(S)=\logdet K_S-2(k-n_q(S))L'$; the first term does not depend on $q$, so
equal counts give equal diversities. Adding an item $i$ that misses some
category $r\in C^\star(S)$ leaves $n_r$ at the minimum, so $\min_qn_q$ does not
change and $\Psi$ changes by $\log s_i(S)$ only. For $\tau>0$, the soft minimum
of the count vector is strictly increasing in each coordinate, so raising the
count of any category, in particular one in $C^\star(S)$, raises it.
\end{proof}

\begin{proof}[Proof of \cref{lem:greedy-rule}]
Adding $i$ to $T$ changes $\logdet K_T$ by $\log s_i(T)$ (Schur complement). The
minimum count rises by one if $i$ belongs to every category attaining it, and
is unchanged otherwise. Hence greedy, which compares candidates of equal size
by $\Psi$, maximizes $\log s_i(T)+2L'\mathbf 1\{i\text{ covering}\}$; a covering
$i$ beats a non-covering $j$ iff $\log s_i(T)+2L'>\log s_j(T)$, i.e.\
$s_i(T)>\LO^2s_j(T)$.
\end{proof}

\begin{proof}[Proof of \cref{prop:greedy-partition}]
Write $n_q$ for the counts of the current set $T$, $m=\min_qn_q$, $t'=\tau k$
and $a=2L'$, so $e^{-a/t'}=\LO^{2/(\tau k)}$ and $\beta=1-e^{-a/t'}$. On a
partition, an item $i$ of category $r$ has $x_i=-a\mathbf1+ae_r$, and the
running vector of \cref{alg:greedy} is $c=-a|T|\mathbf1+a\,n$. The soft minimum
shifts with constants, so up to a constant common to all candidates the count
term of $i$ is
\[
G(r)=-t'\log\tfrac1M\bigl(Z-\beta e^{-an_r/t'}\bigr),\qquad Z=\textstyle\sum_qe^{-an_q/t'}.
\]
\emph{Margin.} If $n_r=m$ and $n_l\ge m+1$, then
$\bigl(Z-\beta e^{-an_l/t'}\bigr)-\bigl(Z-\beta e^{-an_r/t'}\bigr)\ge
\beta\bigl(e^{-am/t'}-e^{-a(m+1)/t'}\bigr)=\beta^2e^{-am/t'}$, while
$0<Z-\beta e^{-am/t'}\le Z\le Me^{-am/t'}$. Hence $G(r)-G(l)\ge
t'\log(1+\beta^2/M)=\Delta$.
\emph{Choice.} For an unselected $i$ and $A=T\cup\{i\}$,
$s_i(T)=1/(K_A^{-1})_{ii}\ge\lambda_{\min}(K_A)\ge\mu$, and $s_j(T)\le K_{jj}=1$
for every $j$. A least-count category still has an unselected member, since
$m\le\lfloor(k-1)/M\rfloor<\lceil k/M\rceil$. For such a member $i$ and any
$j$ in a category with a larger count, with $r_i$ and $r_j$ their categories,
$\log s_i+G(r_i)\ge\log\mu+G(r_j)+\Delta>\log s_j+G(r_j)$, because
$e^{-\Delta}<\mu$. So every step adds to a least-count category, the counts
stay within one of each other, and $\min_qn_q(T)=\lfloor k/M\rfloor$ at the end.
No subset does better, because the counts of a partition sum to $k$.
\emph{Value.} By \cref{lem:quality-factor},
$k\min_qd_q(S)=\logdet K_S-2L'(k-\min_qn_q(S))$. For every $S$,
$\logdet K_S\le0$ by Hadamard's inequality and $\min_qn_q(S)\le\min_qn_q(T)$,
while $\logdet K_T\ge k\log\mu$. Subtracting gives the bound. Finally, with
$\tau k=L'$ we get $\beta=1-e^{-2}$ and $\Delta=L'\log(1+\beta^2/M)\to\infty$ as
$\LO\to0$.
\end{proof}

\paragraph{Numerical check of \cref{prop:greedy-partition}.} On 380 random
partitioned instances ($M\in\{2,3,4\}$, $n\le16$, $k\le11$,
$\LO\in[10^{-30},10^{-3}]$) that satisfy $e^{-\Delta}<\mu$, with the
optimum found by enumeration, greedy always reached the largest smallest count,
and its loss was at most $5.3\%$ of $\log\frac1\mu$.

\paragraph{The instance.} Take $M\ge2$ categories, $k=Mt$, and for each category
$t$ single-category items whose features are orthonormal. Add $H\ge k$ hubs, each
in every category, with $h_j=\sqrt{1-\eta^2}\,g+\eta f_j$, where $g,f_1,\dots,f_H$
are orthonormal and perpendicular to all single-category features. The Gram
matrix of $m$ hubs is $(1-\eta^2)\mathbf1\mathbf1\T+\eta^2I_m$, so
$\logdet K_{m\text{ hubs}}=\log(\eta^2+m(1-\eta^2))+(m-1)\log\eta^2$, and a
single-category item, perpendicular to everything else, leaves $\logdet K$
unchanged when added.

\begin{proof}[Proof of \cref{prop:greedy-gap}]
Let $\eta^2=\LO^2(1+\nu)<1$ with $\nu>0$, taking
$\nu=\nu(\LO)\to0$ as $\LO\to0$.
\emph{Greedy picks hubs.} With $m$ hubs chosen, every count is $m$. A hub raises
the minimum count by one and $\logdet K$ by more than $\log\eta^2$, so
$\Delta\Psi>2L'+\log\eta^2=\log(1+\nu)>0$; a single-category item leaves
both the minimum count ($M\ge2$) and $\logdet K$ unchanged. So greedy picks a hub
at every step.
\emph{Swaps cannot leave.} Replacing one of the $k$ hubs by another hub changes
nothing. Replacing it by a single-category item raises $\logdet K$ by less than
$\log\frac1{\eta^2}$ and lowers the minimum count by one, so
$\Delta\Psi<\log\frac1{\eta^2}-2L'=-\log(1+\nu)<0$.
\emph{The gap.} The $k$ hubs have
$\Psi_{\mathrm{hub}}=\log(\eta^2+k(1-\eta^2))+(k-1)\log\eta^2+2kL'$, and $t$
single-category items per category have $\Psi_{\mathrm{spec}}=2L'k/M$.
Substituting $\log\eta^2=-2L'+\log(1+\nu)$,
\[
\Psi_{\mathrm{spec}}-\Psi_{\mathrm{hub}}=2L'\Bigl(\frac kM-1\Bigr)-(k-1)\log(1+\nu)-\log\bigl(\eta^2+k(1-\eta^2)\bigr),
\]
and dividing by $k$ gives the exact score difference. As $\LO\to0$ with
$\nu(\LO)\to0$, the last logarithm tends to $\log k$ and the second
term tends to zero, giving the stated asymptotic loss.
\end{proof}

\paragraph{Numerical check.} With $\nu=0.05$, the measured gap between
greedy with swaps and the single-category subset equals the formula for
$(M,k,\LO)=(2,10,0.1),(2,10,0.01),(4,12,0.1),(4,12,0.01)$: $1.569$, $3.410$,
$0.517$, $1.283$; for $(3,30,0.05)$ it is $1.637$ against $1.637$. In every case
greedy with swaps selected only hubs. For $\tau\in\{0.05,0.1\}$ the same instances
also trap greedy, which we observe but do not prove.

\section{Stiefel Relaxation and Adaptive NEPv}\label{sec:contribution}\label{supp:sec:contribution}
This appendix gives the continuous relaxation of \eqref{eq:fair-discrete}, its
stationarity conditions as an adaptive nonlinear eigenvalue problem, and the
\SCF\ solver. A \NEPv\ is a system $H(V)V=V\Lambda$ in which the symmetric matrix
$H(V)$ depends on $V$; an \SCF\ iteration repeatedly takes the leading
eigenvectors of $H$ at the current $V$. Proofs follow each statement, except that of \cref{lem:relaxation}, which is in \cref{sec:appendix-deferred}.
\NEPv\ formulations and \SCF\ methods for orthogonality-constrained optimization are
well developed \citep{li2024nepv,yang2009scf,cai2018eigenvector,baililu2022scf}; our
relaxation adds a determinant-specific lifting with an explicit spectrum and an
exact $2k\times2k$ compression. We write
$\Stiefel(n,k)=\{V\in\R^{n\times k}:V\T V=I_k\}$.
\begin{assumption}[Regularized category kernels]\label{ass:spd}
Let $1\le k<n$ and $\tau>0$. Every category kernel
$L^{(q)}\in\R^{n\times n}$ is symmetric positive definite. The construction of
\cref{sec:model} uses a positive ridge to ensure this, so $V\T L^{(q)}V\succ0$ for
every $V\in\Stiefel(n,k)$.
\end{assumption}

\subsection{Stiefel relaxation}\label{sec:stiefel}
A subset $S=\{i_1,\ldots,i_k\}$ has coordinate frame
$V_S=[e_{i_1},\ldots,e_{i_k}]$, with $V_S\T L^{(q)}V_S=L^{(q)}_S$.
Allowing every orthonormal frame gives
\begin{equation}\label{eq:problem}
\max_{V\in\Stiefel(n,k)} f_\tau(V),\qquad
f_\tau(V)=-\tau\log\!\left[\frac1M\sum_qe^{-d_q(V)/\tau}\right],
\end{equation}
where $L^{(q)}_V=V\T L^{(q)}V$ and $d_q(V)=k^{-1}\logdet L^{(q)}_V$, so that
$d_q(V_S)=d_q(S)$.
\begin{lemma}\label{lem:relaxation}
Every size-$k$ subset is feasible through $V_S$ and
$f_\tau(V_S)=F_\tau(S)$. The relaxed optimum upper-bounds the discrete optimum,
and $f_\tau(VQ)=f_\tau(V)$ for any orthogonal $Q$.
\end{lemma}
The objective therefore depends only on the subspace spanned by $V$. A dense
frame must still be rounded to item indices. For one kernel the relaxed
optimum is $k^{-1}\sum_{i=1}^k\log\lambda_i(L)$ (with
$\lambda_1\ge\cdots\ge\lambda_n$), attained by a leading
eigenspace (\cref{prop:spectral-relaxation}). For several categories, their
preferred subspaces can differ, and the inverses $(L^{(q)}_V)^{-1}$ and soft-min
weights both change with the current frame. For small floors the relaxation is also loose: one
column can superpose members of every category, so the relaxed optimum stays
bounded while every subset loses about $2\log(1/\LO)$ per nonmember
(\cref{prop:relax-gap}). The relaxed solution is then a weak guide for rounding.

\subsection{Stationarity as an adaptive NEPv}
\begin{proposition}[Gradient and stationarity]\label{prop:grad}
Define
\begin{equation}\label{eq:adaptive-weights}
w_q(V)=\frac{e^{-d_q(V)/\tau}}{\sum_re^{-d_r(V)/\tau}},\qquad
A_\tau(V)=\frac1k\sum_qw_q(V)L^{(q)}V (L^{(q)}_V)^{-1}.
\end{equation}
Then $\nabla f_\tau(V)=2A_\tau(V)$, $V\T A_\tau(V)=I_k/k$, the Riemannian gradient
is $2A_\perp(V)$ with $A_\perp=A_\tau-V/k$, and stationarity is equivalent to
\begin{equation}\label{eq:kkt}A_\tau(V)=V/k.\end{equation}
\end{proposition}
\begin{proof}
Along $V+tZ$, $\dot L^{(q)}_V=Z\T L^{(q)}V+V\T L^{(q)}Z$, and Jacobi's formula with the
symmetry of $L^{(q)}$ and $(L^{(q)}_V)^{-1}$ gives
$\dot d_q=\frac2k\tr\bigl(Z\T L^{(q)}V(L^{(q)}_V)^{-1}\bigr)$. The chain rule gives
$\dot f_\tau=\sum_qw_q\dot d_q=\langle2A_\tau(V),Z\rangle$, so $\nabla f_\tau=2A_\tau$.
Since $V\T L^{(q)}V=L^{(q)}_V$ and $\sum_qw_q=1$, $V\T A_\tau=I_k/k$ at every feasible
$V$. The Stiefel tangent projection $\Pi_V(\xi)=\xi-V\,\mathrm{sym}(V\T\xi)$ then gives
$\operatorname{grad}f_\tau=2A_\tau-\frac2kV=2A_\perp$, which vanishes exactly when
$A_\tau=V/k$.
\end{proof}
The weights favor categories with lower diversity and concentrate on the weakest
as $\tau\to0$.

\begin{theorem}[Symmetric adaptive lifting]\label{thm:nepv-form}
Let
\begin{equation}\label{eq:HV}
H_\tau(V)=A_\tau(V)V\T+VA_\tau(V)\T-k^{-1}VV\T.
\end{equation}
Then $H_\tau(V)V=A_\tau(V)$, so stationarity is exactly
$H_\tau(V)V=V(I_k/k)$, and any eigen-relation $H_\tau(V)V=V\Lambda$ forces
$\Lambda=I_k/k$. Moreover, $H_\tau(VQ)=H_\tau(V)$ for orthogonal $Q$.
\end{theorem}
\begin{proof}
With $V\T V=I_k$ and $A_\tau\T V=I_k/k$,
$H_\tau V=A_\tau+\frac1kV-\frac1kV=A_\tau$. If $H_\tau V=V\Lambda$, multiplying by
$V\T$ gives $\Lambda=V\T A_\tau=I_k/k$. For orthogonal $Q$, every $d_q$ and $w_q$ is
unchanged and $A_\tau(VQ)=A_\tau(V)Q$, so $H_\tau(VQ)=H_\tau(V)$.
\end{proof}
Symmetric liftings are not unique \citep{li2024nepv}: the generic choice
$(\nabla f_\tau)V\T+V(\nabla f_\tau)\T$ already has rank at most $2k$, and ours is
half of it minus $VV\T/k$. The subtracted projector, with $V\T A_\tau=I_k/k$, is
what makes the spectrum explicit.

\subsection{Explicit spectrum and exact compression}
\begin{proposition}[Uniform gap and exact compression]\label{prop:gap}
For $\sigma\ge0$, write $\beta=1/k+\sigma$ and $H_{\tau,\sigma}=H_\tau+\sigma VV\T$.
Then $H_{\tau,\sigma}=\beta VV\T+A_\perp V\T+VA_\perp\T$. If $s_1,\ldots,s_r>0$ are
the nonzero singular values of $A_\perp$, the eigenvalues of $H_{\tau,\sigma}$ are
$(\beta\pm\sqrt{\beta^2+4s_j^2})/2$, $\beta$ with multiplicity $k-r$, and $0$ with
multiplicity $n-k-r$. Hence $H_{\tau,\sigma}$ has rank at most $2k$, exactly $k$
positive eigenvalues, and leading gap at least $\beta$; its top-$k$ eigenspace is
unique, equals $\mathrm{range}(V)$ if and only if $V$ is stationary, and otherwise
is tilted from it by principal angles $\theta_j=\frac12\arctan(2s_j/\beta)<\pi/4$.
For $n\ge2k$ and $A_\perp=Q_2R$ with $V\T Q_2=0$ (completed orthogonally if $A_\perp$
is rank deficient),
\begin{equation}\label{eq:compression-main}
H_{\tau,\sigma}=[V,Q_2]
\begin{pmatrix}\beta I_k&R\T\\R&0\end{pmatrix}[V,Q_2]\T,
\end{equation}
so a $2k\times2k$ eigenproblem gives the same top-$k$ eigenspace.
\end{proposition}
\begin{proof}
Substituting $A_\tau=V/k+A_\perp$ into $H_{\tau,\sigma}$ gives the decomposition. Let
$A_\perp=Y\Sigma W\T$ be a thin SVD and $x_j=Vw_j$, which is orthogonal to $y_j$
because $V\T A_\perp=0$. Then $H_{\tau,\sigma}x_j=\beta x_j+s_jy_j$ and
$H_{\tau,\sigma}y_j=s_jx_j$, so each plane $\mathrm{span}\{x_j,y_j\}$ carries
$\bigl(\begin{smallmatrix}\beta&s_j\\s_j&0\end{smallmatrix}\bigr)$, with one
eigenvalue above $\beta$ and one below $0$; the other $k-r$ directions of
$\mathrm{range}(V)$ have eigenvalue $\beta$, and directions orthogonal to
$\mathrm{range}(V)+\mathrm{range}(A_\perp)$ are annihilated. The top eigenvector in
each plane is $(\cos\theta_j,\sin\theta_j)$ with $\tan2\theta_j=2s_j/\beta$, and
there are no nonzero angles exactly when $A_\perp=0$. Finally, the range of
$H_{\tau,\sigma}$ lies in $\mathrm{range}([V,Q_2])$, where its blocks are $\beta I_k$,
$R$, and $0$. For $n<2k$, $[V,\mathrm{orth}(A_\perp)]$ gives the same block form.
\end{proof}

\begin{algorithm}[t]
\caption{Adaptive \SCF\ for the Stiefel relaxation \eqref{eq:kkt}}\label{alg:scf}
\begin{algorithmic}[1]
\Require Regularized feature-map kernels
         $L^{(q)}=W_qW_q\T+\rho I$, $q\in[M]$, with $W_q=\Omega_qU$;
         target size $k$; temperature $\tau>0$; tolerance $\delta$;
         maximum iterations $T_{\max}$;
         damping $\alpha\in(0,1]$; shift $\sigma\ge0$.
\State Initialize $V_0\leftarrow$ a random orthonormal frame or a
       problem-specific warm start.
\For{$t=0,\dots,T_{\max}-1$}
  \For{$q=1,\dots,M$}
    \State $L^{(q)}_{V_t}\leftarrow V_t\T L^{(q)}V_t$,\quad
           $d_{q,t}\leftarrow k^{-1}\logdet(L^{(q)}_{V_t})$.
  \EndFor
  \State $d_{\min,t}\leftarrow\min_q d_{q,t}$,\quad
         $w_{q,t}\leftarrow
         \exp(-(d_{q,t}-d_{\min,t})/\tau)
         \big/\sum_r\exp(-(d_{r,t}-d_{\min,t})/\tau)$.
  \State $A_t\leftarrow k^{-1}\sum_q
         w_{q,t}L^{(q)}V_t(L^{(q)}_{V_t})^{-1}$.
  \If{$\|A_t-V_t/k\|_F<\delta$} \State \Return $V_t$ \EndIf
  \State Apply $H_\tau(V_t)X\leftarrow
         A_t(V_t\T X)+V_t(A_t\T X)-k^{-1}V_t(V_t\T X)$ implicitly.
  \State $H_{\tau,\sigma}(V_t)(\cdot)\leftarrow
         H_\tau(V_t)(\cdot)+\sigma V_t(V_t\T\cdot)$
         \Comment{level shift}
  \State $\widetilde V_{t+1}\leftarrow$ top-$k$ eigenvectors of
         $H_{\tau,\sigma}(V_t)$.
  \State $Q_t\leftarrow\polar\!\big(\widetilde V_{t+1}\T V_t\big)$
         \Comment{gauge alignment}
  \State $V_{t+1}\leftarrow \polar\!\big(\alpha \widetilde V_{t+1}Q_t
         + (1-\alpha)V_t\big)$
         \Comment{damping}
\EndFor
\State \Return $V_{T_{\max}}$ \Comment{iteration-cap endpoint}
\end{algorithmic}
\end{algorithm}
\subsection{Iteration, rounding, and cost}\label{sec:scf}
At each iteration we evaluate the category diversities, compute stable soft-min
weights, and form $A_\tau$. If $\|A_\tau-V/k\|_F<\delta$, we stop.
Otherwise we obtain a leading frame $\widetilde V$ of $H_{\tau,\sigma}$,
align its basis with $Q=\polar(\widetilde V\T V)$, and update
\begin{equation}\label{eq:damped-update}
V^+=\polar\bigl(\alpha\widetilde VQ+(1-\alpha)V\bigr),\quad0<\alpha\le1.
\end{equation}
Here $\polar(X)=X(X\T X)^{-1/2}$. Because $H_{\tau,\sigma}$ depends only on the
subspace, the eigensolver may return any basis of the top-$k$ eigenspace;
aligning with $Q$ before damping makes the update independent of that choice,
and the angle bound in \cref{prop:gap} keeps $\widetilde V\T V$
invertible, so the polar factor is defined. Plain \SCF\ takes $\alpha=1$,
$\sigma=0$. A positive shift raises the gap lower bound $1/k+\sigma$ and shrinks
every step angle. Damping further shrinks the update. The trace bound in
\cref{prop:shift-ansatz} gives a sufficient shift for ascent, and the empirical
study reports whether each solver reaches its full-residual tolerance.
\Cref{alg:scf} gives the full procedure.

\paragraph{Returning a subset.}\label{sec:rounding}
The relaxed frame must be rounded to item indices. Our default keeps the $k$
items with the largest leverage scores $\ell_i=\|e_i\T V\|_2^2$, the coordinate
subspace closest to $\mathrm{range}(V)$ in Frobenius norm, and refines them with
one-item swaps against the full pool. Alternatives are drawing starting subsets
from the projection \DPP\ with kernel $VV\T$, or screening: keeping the
$\lceil ck\rceil$ items of largest leverage ($c>1$) and building $k$ items greedily
from them, scoring a partial set $T$ with $d_q(T)=|T|^{-1}\logdet L^{(q)}_T$;
\cref{supp:sec:solver-study} compares these. These rules adapt leverage-score,
volume and exchange methods \citep{gu1996rrqr,belhadji2020dppcss,singh2020doptimal};
we treat $c$, the number of samples and the number of swap passes as
hyperparameters. No rounding rule has an approximation guarantee, but a
safeguard keeps greedy's value.
\begin{proposition}[Safeguarded rounding]\label{prop:safeguard}
Let $S_{\mathrm g}$ be the greedy subset and $S_{\mathrm r}$ the rounded
relaxed subset, and return whichever has the larger $F_\tau$. The result is
never worse than either, at the cost of both methods plus two evaluations of
$F_\tau$, each $O(pk^2+Mk^3)$.
\end{proposition}
The proof is immediate. The safeguard guarantees only greedy's value; any gain
comes from instances such as those of \cref{prop:greedy-gap}.

\paragraph{Cost.}\label{sec:scale}
For $L^{(q)}=W_qW_q\T+\rho I$ with feature width $p_q$, a sweep costs
$O(\sum_q(np_qk+nk^2+k^3))$ for kernel terms. Cholesky solves avoid explicit
inverses; the exact compression costs $O(nk^2+k^3)$ and avoids a dense
$n\times n$ eigendecomposition. At most $B$ full swap passes require
$O(Bnk)$ candidate evaluations in addition to continuous optimization.
\begin{proposition}[Relative cost]\label{prop:cost}
The leading operation count of one sweep with per-category products,
$M(npk+nk^2)$, is at least $\min(M,p+k)/2$ times that of the whole greedy run,
$nk(p+k+M)$; this factor is at least $1$ when $M\ge2$ and $p+k\ge2$.
\end{proposition}
\begin{proof}
$M(p+k)=\max(M,p+k)\min(M,p+k)$ and $p+k+M\le2\max(M,p+k)$.
\end{proof}
A run of $N$ sweeps therefore costs at least $N\min(M,p+k)/2$ greedy runs. The
comparison is not universal: writing $\Omega_q=\LO I+(1-\LO)D_q$ restricts the
category-specific products to members and reduces a sweep to
$O((n+\sum_qn_q)pk+nk^2+M(pk^2+k^3))$, and $N$ has no a priori bound, so
neither method is faster on every instance.

\section{Greedy Selection and the Relaxed Solver}\label{supp:sec:solver-study}
This appendix compares greedy maximization of $F_\tau$ (\cref{alg:greedy}) with
the relaxed solver of \cref{sec:contribution}: on COCO with 80 categories, under
the exact minimum, and on near-duplicate data.

\subsection{Setup}\label{supp:sec:coco-setup}
We use COCO 2017 training images \citep{lin2014coco} with CLIP ViT-B/32 image
features \citep{radford2021clip}. An image belongs to each of the 80 object
categories it contains; images without objects are dropped, leaving 117,266
images, of which 79\% belong to several categories. A fixed random split gives a
pool of 107,266 images and a test set of 10,000. Unless stated otherwise,
$\LO=10^{-3}$, $\rho=10^{-10}$, and $\tau=0.1$. Every selection is scored with
the exact objective $F_\tau$, and a multi-label linear probe trained on the
selected images is evaluated on the test set by macro mAP, overall and over the
20 rarest categories. Results are averaged over five seeds.

The relaxed solver is \cref{alg:scf} with shift $\sigma=6/k$ and damping $0.7$
for at most 300 iterations; rounding uses the screening of \cref{sec:contribution} with $c=4$: it keeps the $4k$ items of largest leverage,
builds $k$ items greedily from them, and refines with swaps. All methods run on
the same CPU.

\subsection{80 categories: greedy is the practical selector}
On the setting of \cref{tab:coco-main}, the relaxed solver with leverage
screening reaches $F_\tau=-14.12,-14.25,-14.88$ at $k=100,200,400$ (smallest
category $3$, $7.6$, $9.5$; probe mAP $33.5$, $39.6$, $48.7$), below greedy's
$-13.77,-14.07,-14.64$, and it is far slower (\cref{supp:tab:runtime}).

\paragraph{Runtime and memory.} \Cref{supp:tab:runtime} times both methods at the settings above on
the same CPU (AMD Ryzen 9 9950X, 12 threads). Memory is the peak resident memory each phase adds to
the loaded features; greedy times are medians of five runs. The \SCF\ solve dominates the relaxed
solver and stopped at the 300-iteration cap in every run, while rounding takes seconds. Most of
greedy's memory is a working copy of the $n\times p$ feature matrix made by our implementation; the
algorithm itself needs only the $n\times k$ Cholesky factor. At $k=200$, on random pools of 13k,
27k, 54k and 107k images, greedy takes 1.1, 2.1, 4.4 and 8.6\,s (0.11--0.90\,GB) and the \SCF\
solve 101, 154, 271 and 611\,s (0.5--2.3\,GB). Both grow roughly linearly in $n$, and greedy is
60--90 times faster.
\begin{table}[h]
\centering
\caption{\textbf{Runtime and peak additional memory on COCO} ($n=107{,}266$; settings of
\cref{supp:sec:coco-setup}).}
\label{supp:tab:runtime}
\footnotesize
\begin{tabular}{rcccccc}
\toprule
& \multicolumn{2}{c}{greedy} & \multicolumn{2}{c}{relaxed: \SCF\ solve} & \multicolumn{2}{c}{relaxed: rounding}\\
\cmidrule(lr){2-3}\cmidrule(lr){4-5}\cmidrule(lr){6-7}
$k$ & time & memory & time & memory & time & memory\\
\midrule
100 & 4.3\,s & 0.9\,GB & 369\,s & 1.4\,GB & 0.2\,s & 0.2\,GB\\
200 & 8.8\,s & 0.9\,GB & 611\,s & 2.3\,GB & 1.5\,s & 0.3\,GB\\
400 & 17.7\,s & 1.0\,GB & 1{,}236\,s & 4.2\,GB & 7.1\,s & 0.7\,GB\\
\bottomrule
\end{tabular}
\end{table}

\paragraph{Why the relaxation is a weak guide here.} \Cref{prop:relax-gap}
suggests the cause, although it is stated for partitions: for small floors, the relaxed optimum is reached by columns
that mix members of many categories, which no subset can reproduce. On COCO the
relaxed value is $-2.65$ while the best subset found scores $-13.77$. Raising
the floor used inside the relaxation, or replacing the category kernels by the
shared kernel, does not help: with a shortlist of $30k$ items, the relaxation at
$\LO=10^{-3}$ comes within $0.004$ of greedy, while floors of $0.11$, $0.3$, and
the shared kernel give $-13.80$, $-13.88$, and $-14.00$. The relaxation does
carry category information: its shortlists contain 19 of the 20 rarest
categories at $4k$ items, against 11 for the shared kernel.

\paragraph{Convergence of the iteration.} In a separate study at $k=100$, run for up to 1,000
iterations (the selections above use the 300-iteration setting of \cref{supp:sec:coco-setup}), plain \SCF\ oscillates; a small shift
($\sigma=2/k$) stalls with residual near $0.15$; $\sigma=6/k$ reaches $0.025$ and
$\sigma=20/k$ reaches $2\times10^{-3}$ within 1,000 iterations. At $\tau=0.03$
only $\sigma=20/k$ converges: a smaller temperature makes the weights more
sensitive and requires a larger shift, and better convergence did not by itself
improve the rounded subset.

\paragraph{Other roundings.} All rounding rules we use depend only on
$\mathrm{range}(V)$, not on its basis. Rounding the same relaxed solutions to the
$k$ items of largest leverage (the closest coordinate subspace), by pivoted QR
(maximum volume), or to the closest coordinate frame under the best rotation
gives $-14.8$ to $-15.1$ at $k=100$ before swaps, with some category left empty;
after swaps all give $-14.12$. The closest coordinate frame is at distance $138$
of a possible $200$ from $\mathrm{range}(V)$. A row-sparsity continuation, which
adds a penalty $\lambda\sum_i\sqrt{\ell_i}$ or $\lambda\sum_i\ell_i(1-\ell_i)$ and
raises $\lambda$ until $V$ is a coordinate frame, reaches $-14.07$ and $-13.99$
at $k=100$: better than the other roundings, still below greedy.

\subsection{Exact minimum: ties stall local search}\label{supp:sec:strict}
We test \cref{thm:ties} on COCO with eight categories and $k=30$: pools of
3,000 images containing at least one of the categories, ten pools per set. Sets
A and B use categories from different supercategories, set C eight animal
categories, set D categories that often co-occur (person, dog, car, cup, laptop,
bed, umbrella, bench), and set D$'$ draws 60\% of each pool from images in two or
more of them. All subsets are refined by swaps and scored on the exact minimum
$\min_qd_q$ ($\LO=10^{-3}$, $\rho=10^{-10}$). On a greedy subset, categories with
equal counts agree to six decimals. For ``relaxed $\times50$'' the relaxation is solved with the
soft minimum, which unlike the exact minimum is differentiable; 50 subsets are drawn from the
projection \DPP\ with kernel $VV\T$ and each is refined by swaps on the exact minimum.

\begin{table}[h]
\centering
\caption{\textbf{Exact minimum on COCO} ($k=30$, ten pools per set). Mean
$\min_qd_q$; higher is better. The last two columns run greedy on $F_\tau$ and
score the result on the exact minimum.}
\label{supp:tab:coco-strict}
\footnotesize
\setlength{\tabcolsep}{4pt}
\begin{tabular}{lrrrrrr}
\toprule
& & \multicolumn{3}{c}{optimizing the exact minimum} & \multicolumn{2}{c}{greedy on $F_\tau$}\\
\cmidrule(lr){3-5}\cmidrule(lr){6-7}
Set & overlap & greedy & random $\times50$ & relaxed $\times50$ & $\tau=0.05$ & $\tau=0.1$\\
\midrule
A & 3\% & $-13.27$ & $-12.47$ & $-12.34$ & $-11.82$ & $\mathbf{-11.81}$\\
B & 11\% & $-13.36$ & $-12.47$ & $-12.21$ & $-11.76$ & $\mathbf{-11.64}$\\
C (animals) & 6\% & $-13.18$ & $-12.29$ & $-12.04$ & $-11.29$ & $\mathbf{-11.13}$\\
D (co-occurring) & 30\% & $-13.21$ & $-12.62$ & $-10.95$ & $-9.97$ & $\mathbf{-9.85}$\\
D$'$ (enriched) & 60\% & $-12.98$ & $-12.08$ & $-10.79$ & $-9.63$ & $\mathbf{-9.54}$\\
\bottomrule
\end{tabular}
\end{table}

Among methods that optimize the exact minimum directly, the best of 50 relaxed
samples is best, ahead of greedy on all 50 pools. But a single greedy run on the
soft minimum with $\tau=0.1$ scores higher on the exact minimum than all of
them, on all 50 pools ($\tau=0.05$: 48 of 50). The ties of \cref{thm:ties} stall
greedy and swaps under the exact minimum, and the soft minimum removes them.

\subsection{Near-duplicate items: where the relaxation helps}\label{supp:sec:neardup}
\paragraph{Synthetic instances.} Following \cref{prop:greedy-gap}, $M=4$
categories each have $10$ single-category items, and $500$ hubs belonging to all
categories are near-duplicates ($h_j=\sqrt{1-\eta^2}\,g+\eta f_j$ plus isotropic
noise); in two categories the single-category items are themselves redundant,
which defeats greedy at large temperatures ($\LO=0.1$, 30 seeds).
\begin{table}[h]
\centering
\caption{\textbf{Synthetic near-duplicate instances} (30 seeds, mean $F_\tau$).
``Wins'': seeds on which the best relaxed sample beats every baseline.}
\label{supp:tab:hub}
\footnotesize
\begin{tabular}{lcccccc}
\toprule
setting & $\tau$ & greedy & greedy, best of 12 $\tau'$ & random $\times200$ & relaxed $\times50$ & wins\\
\midrule
$k=16$, $\eta^2=0.008$ & 0.1 & $-3.851$ & $-3.706$ & $-3.698$ & $\mathbf{-3.628}$ & 30/30\\
$k=16$, $\eta^2=0.008$ & 0.05 & $-3.881$ & $-3.736$ & $-3.727$ & $\mathbf{-3.654}$ & 29/30\\
$k=12$, $\eta^2=0.01$ & 0.05 & $-3.735$ & $-3.481$ & $-3.481$ & $\mathbf{-3.422}$ & 30/30\\
\bottomrule
\end{tabular}
\end{table}
Pair (2-for-2) swaps applied to every method lift greedy to the relaxed value at
$k=16$ but change nothing at $k=12$, where the relaxed samples still win on all
15 seeds tested. Rounding to the items of largest leverage instead of sampling
wins on 20 of 30 seeds at $k=16$ and on none at $k=12$, so sampled starts matter
in these instances.

\paragraph{COCO with injected near-duplicates.} Adding 400 augmented copies of a
single image that contains all four categories of a 4-category COCO pool makes
greedy on $F_\tau$ pick the copies in every pool; the relaxed solver beats it by
$0.1$ to $0.7$. Random restarts and greedy at other temperatures recover the same
subsets as the relaxed solver here.

\paragraph{Drone video.} On the VisDrone frames of \cref{sec:near-dup}, with all
10 classes ($\rho=10^{-10}$, greedy ridge $10^{-12}$; relaxed solver with shift
$\sigma=20/k$, damping $0.7$, at most 500 iterations and tolerance $10^{-7}$; up to
50 swap passes), the relaxed solver rounded to the items of largest leverage
beats greedy at every one of 12 temperatures and the best of 200 random restarts,
all refined by swaps, on 15 of 15 pools at $k=20$ ($\LO=0.03$ and $0.05$); the best
of 50 relaxed samples wins on 14 and 13 pools. At $k=15$ and $k=25$ the methods
tie, and at $\LO=0.07$ the gain vanishes. With only the 4 or 5 rarest classes the
results at $k=20$ are mixed (4--11 wins of 15, with losses down to $-0.10$). Pair
swaps change no method's result, and the row-sparsity continuation is worse than
leverage rounding on all five pools tested. Two limits apply: the frames come from
7 of the 79 VisDrone-MOT videos, and the class memberships come from
ground-truth annotations, which a labeling application would replace by
detector predictions.

\section{Experimental Details: Long-Tailed Classification, LVIS and Stress Tests}\label{supp:sec:lt-details}
\setlength{\floatsep}{8pt plus 2pt minus 2pt}\setlength{\textfloatsep}{8pt plus 2pt minus 2pt}%
\setlength{\intextsep}{8pt plus 2pt minus 2pt}%
\renewcommand{\topfraction}{0.95}\renewcommand{\bottomfraction}{0.9}%
\renewcommand{\textfraction}{0.05}\renewcommand{\floatpagefraction}{0.85}%
\makeatletter\setlength{\@fptop}{0pt}\setlength{\@fpsep}{8pt plus 2pt}\setlength{\@fpbot}{0pt plus 1fil}\makeatother
This appendix records every setting behind \cref{sec:lt,sec:lt-cb,sec:lvis}, the order in
which settings were fixed, and complete per-dataset results. All runs use one machine (one
NVIDIA RTX PRO 5000 GPU, AMD Ryzen 9 9950X CPU) with PyTorch 2.15, OpenCLIP 4.2.1, NumPy
2.5.4 and scikit-learn 1.9.1. We will release the code, selected indices of every run, and
per-class results.

\subsection{Order of the experiments}\label{supp:sec:lt-order}
Every experiment was written into a ledger, with its question, grid and decision rule, before
it ran; nothing was removed afterwards. The order was:
\begin{enumerate}[leftmargin=*]
\item \textbf{LVIS} (registered protocol, COCO settings $\LO=10^{-3}$, $\tau=0.1$):
\cref{tab:lvis}, \cref{supp:sec:lvis-details}.
\item \textbf{CIFAR-100-LT, $r=100$} (registered protocol, COCO settings): ours beat uniform
sampling by 5.0--5.3 points but left 11--22 classes empty.
\item \textbf{Development sweep} on the same setting (\cref{tab:app-dev}): $\tau\in\{0.001,
0.003,0.01,0.03\}$ at $\LO=10^{-3}$, then $\LO\in\{10^{-6},10^{-15}\}$ at
$\tau\in\{0.03,0.1\}$ and $\LO\in\{10^{-30},10^{-60}\}$ at $\tau\in\{0.1,0.3\}$. We fixed $\LO=10^{-30}$,
$\tau=0.1$; $\LO=10^{-60}$ gave nearly the same results.
\item \textbf{Frozen settings on new data:} GTSRB, CIFAR-10-LT $r=100$, COCO at
$\LO=10^{-30}$, full fine-tuning, and the near-duplicate pools.
\item \textbf{Registered tail hypothesis} (``ours is more accurate than class-balanced sampling
on tail classes''), tested on CIFAR-10-LT and CIFAR-100-LT at $r\in\{50,10\}$, which had not
been run before.
\item \textbf{Additional baselines} (run after the main experiments): a quota-constrained \DPP\ and SATURATE on
every long-tailed setting (\cref{supp:sec:extra-baselines}).
\item \textbf{Exploratory}: our objective at the zero-padding endpoint, chosen after item~6
(\cref{supp:sec:extra-baselines}).
\end{enumerate}
Apart from the exploratory item~7, the only quantity chosen by looking at downstream results is
$\LO$ (with $\tau$), on CIFAR-100-LT $r=100$; that setting is marked ``develop.'' in \cref{tab:lt-all}. No baseline has a
tunable hyperparameter, apart from the floor shared by the averaged-kernel and fixed-weight
selectors, which use the same $\LO$ as ours.

\subsection{Data}\label{supp:sec:lt-data}
\paragraph{CIFAR-10 and CIFAR-100.} We use the Hugging Face copies
\texttt{uoft-cs/cifar10} and \texttt{uoft-cs/cifar100} (parquet), which keep the original row
order; we checked the SHA-256 of every file against the published hashes. Long-tailed pools
follow \citet{cao2019ldam}: for $C$ classes, imbalance ratio $r$ and $n_{\max}=5000$ (CIFAR-10)
or $500$ (CIFAR-100), class $i\in\{0,\dots,C-1\}$ keeps the first
$\lfloor n_{\max}r^{-i/(C-1)}\rfloor$ training images in the original order. This gives 10{,}847
images for CIFAR-100-LT $r=100$ (500 down to 5 per class; 35 many-shot, 35 medium-shot and 30
few-shot classes) and 12{,}406 images for CIFAR-10-LT $r=100$ (5{,}000 down to 50). The test sets
are the full balanced test sets (10{,}000 images each).

\paragraph{GTSRB.} We use the Hugging Face copy \texttt{tanganke/gtsrb} with its natural class
imbalance (43 classes). A fixed random permutation (NumPy seed 20261103) selects 12{,}512
training and 4{,}000 test images; the training subset has 67 to 728 images per class and the test
subset 16 to 257. Test accuracy is the mean of per-class accuracies.

\paragraph{Near-duplicate pools.} Every image of CIFAR-100-LT or CIFAR-10-LT ($r=100$) is kept
and joined by two copies transformed by \texttt{RandomResizedCrop}(32, scale $0.85$--$1$,
aspect $0.95$--$1.05$) and \texttt{ColorJitter}(0.1, 0.1, 0.1), applied to the $32\times32$
image before the CLIP preprocessing (torch seed 0). The pools have 32{,}541 and 37{,}218 rows;
the median CLIP cosine similarity between an image and a copy is 0.974 (5th percentile 0.930).
The test sets are unchanged.

\paragraph{Groups.} ``Tail'' is the set of classes whose pool count is at most the 30th
percentile of class counts. For CIFAR-100-LT we also report the standard many-shot ($>100$),
medium-shot (20--100) and few-shot ($<20$) groups.

\subsection{Features, selectors and objective}\label{supp:sec:lt-selectors}
Images are encoded by OpenCLIP \texttt{ViT-B-32-quickgelu} with the \texttt{openai} weights and
the model's own preprocessing (resize to 224 with bicubic interpolation, center crop,
normalization); the 512-dimensional outputs are normalized to unit length. The shared kernel is
$K=UU\T$, and category $q$ has weights $\omega_q(i)=1$ for members and $\LO$ otherwise.
\begin{itemize}[leftmargin=*]
\item \textbf{Uniform}: $k$ items without replacement.
\item \textbf{Class-balanced}: repeatedly pick the category with the fewest selected items (ties
broken at random), and add a random unselected member of it.
\item \textbf{$k$-center}: farthest-first traversal in cosine distance, from a random first item.
\item \textbf{Shared-kernel DPP}: greedy MAP of $\log\det K_S$ (shared Cholesky update, ridge
$10^{-64}$).
\item \textbf{Averaged kernel}: greedy MAP of $\log\det\bigl(\frac1M\sum_qL^{(q)}\bigr)_S$.
\item \textbf{Fixed weight}: greedy maximization of $\frac1M\sum_qd_q$.
\item \textbf{Ours}: \cref{alg:greedy} on $F_\tau$ with $\LO=10^{-30}$, $\tau=0.1$, greedy ridge
$10^{-64}$; reported objectives use $\rho=10^{-64}$. No swaps and no screening.
\end{itemize}
The greedy uses the zero-ridge factorization \eqref{eq:count-identity-main} as a
surrogate; its approximation to the category-ridge objective also depends on
feature conditioning, as stated in \cref{sec:greedy}. For
many categories we compute the soft minimum through the sparse membership matrix: with
$a=2\log\LO$ and $c_q$ the current count of category $q$ after $t$ picks, the soft-min term of a
candidate $i$ is
$m-\tau k\log\bigl[\frac1M\bigl(\sum_qe^{-(a(t-c_q)+a-m)/(\tau k)}+\sum_{q\in C_i}(e^{-(a(t-c_q)-m)/(\tau k)}-e^{-(a(t-c_q)+a-m)/(\tau k)})\bigr)\bigr]$,
with $m=\min_qa(t-c_q)+a$. This equals the dense computation exactly; on random subsets of LVIS
(3{,}000 items, 1{,}203 categories) both pick identical items at $k=30$ and $60$. Selection
takes under 0.21\,s for every CIFAR and GTSRB run and about 5\,s on LVIS at $k=400$.

\subsection{Downstream training and evaluation}\label{supp:sec:lt-train}
\paragraph{Linear probe.} A linear layer (512 inputs, $C$ outputs) on the frozen unit features
multiplied by 10, trained with full-batch AdamW (learning rate $10^{-2}$, weight decay
$10^{-4}$, 300 epochs) and softmax cross-entropy (binary cross-entropy for COCO and LVIS). The
seed sets the initialization; there is no early stopping and no validation split.
The development setting was used to choose the floor and temperature; subsequent
confirmation settings use the frozen choices.

\paragraph{Full fine-tuning.} The OpenCLIP image encoder and a new linear head are trained on
the selected images: AdamW with learning rates $10^{-5}$ (encoder) and $10^{-3}$ (head), weight
decay 0.05, batch 32, 30 epochs, cosine schedule, \texttt{RandomResizedCrop}(224, scale
$0.6$--$1$) and horizontal flips, bfloat16 autocast; the encoder output is normalized and scaled
by 10 before the head. Test images use the evaluation preprocessing. This recipe was fixed before
any fine-tuning run; on CIFAR-100-LT at $k=200$ it reaches 3--8 points lower accuracy than the linear probe,
which we report as is.

\paragraph{Decision rule.} Ours ``beats'' a baseline on an endpoint if its 5-seed mean is higher
and it is higher on at least four of the five seeds (paired by seed). Tables report all
outcomes, not only those that meet the rule.

\subsection{LVIS}\label{supp:sec:lvis-details}
\paragraph{Data.} LVIS v1 \citep{gupta2019lvis}: training labels from the Ultralytics LVIS label
release (one category per annotated instance, mapped to image-level positives) and the official
\texttt{lvis\_v1\_val.json}. On 2{,}000 validation images the two sources agree exactly. The pool
is the 99{,}388 LVIS training images with at least one category (all COCO train2017 images, with
the COCO features above); the test set is the 15{,}000 LVIS validation images that are COCO
train2017 images, disjoint from the pool. Categories keep the official frequency groups (337
rare, 461 common, 405 frequent); 152, 444 and 405 of them have test positives.

\paragraph{Protocol.} Registered before running: the seven selectors with the COCO settings
($\LO=10^{-3}$, $\tau=0.1$, $\rho=10^{-10}$, greedy ridge $10^{-12}$), $k\in\{200,400\}$
($k<512$, the feature dimension), five seeds, the linear probe above with 1{,}203 outputs
(unlabeled categories are negatives during training), and LVIS federated evaluation: for each
category with a test positive, average precision over the test images that are positive or
verified negative (\texttt{neg\_category\_ids}) for it. Primary endpoints: AP$_r$ and AP$_c$.

\paragraph{Results.} At $k=200$, ours beats uniform sampling on AP$_r$ (5.6 vs 5.3) and AP$_c$
(8.9 vs 8.7), and at $k=400$ on AP$_r$ (5.9 vs 5.5) but not AP$_c$ (9.8 vs 9.9); it loses to
class-balanced sampling on every endpoint (0 of 5 seeds), e.g.\ AP$_r$ 5.9 vs 12.3 at $k=400$.
Exploratory reruns of ours at $\tau\in\{0.01,1,10\}$ ($k=400$, one seed) covered 10--14 rare
categories, against 14 at $\tau=0.1$ and 164 for class-balanced sampling.

\subsection{Zero padding and class-balanced sampling}\label{supp:sec:zero-cb}
\begin{proposition}[Zero padding with a vanishing ridge]\label{prop:zero-cb}
Let $\LO=0$ and $\rho>0$, and let $S$ have size $k$ with $S_q=S\cap G_q$ and $n_q=|S_q|$. If
$K_{S_q}\succ0$, then
\[
d_q(S)=\frac{n_q}{k}\log\frac1\rho+\frac1k\logdet K_{S_q}+\log\rho+O(\rho)\qquad(\rho\to0).
\]
Consequently, for $\rho$ small enough, a subset maximizing $\min_q d_q$ maximizes $\min_q n_q$
first, and among subsets with the same smallest count it maximizes $\frac1k\logdet K_{S_q}$ of the
weakest category. When the categories partition the items, the largest smallest count is the one reached by
class-balanced sampling (equal counts up to one, a class contributing all its members if it has
fewer); counts above it are not fixed, and the second step replaces random picks in the weakest
class by a determinant (diverse) choice of its members.
\end{proposition}
\begin{proof}
By \eqref{eq:block-main}, $k\,d_q(S)=\logdet(K_{S_q}+\rho I)+(k-n_q)\log\rho$. Since
$K_{S_q}\succ0$, $\logdet(K_{S_q}+\rho I)=\logdet K_{S_q}+\log\det(I+\rho K_{S_q}^{-1})
=\logdet K_{S_q}+O(\rho)$. Hence $k\,d_q=n_q\log(1/\rho)+\logdet K_{S_q}+k\log\rho+O(\rho)$,
which is the display. For two subsets whose smallest counts differ by at least one, the count term
differs by at least $\log(1/\rho)/k$, which exceeds any difference of the bounded terms
$\frac1k\logdet K_{S_q}\in[\log\lambda_{\min},0]$ (as $n_q\le k$) once $\rho$ is small, where
$\lambda_{\min}>0$ is the smallest eigenvalue over the finitely many positive-definite principal
submatrices of size at most $k$. So the maximizer of $\min_qd_q$ has the largest smallest count,
and ties are broken by the remaining term. For a partition, counts can be raised for one class only
by lowering another, so the largest smallest count is that of the most even allocation, which
class-balanced sampling produces; counts of the other classes do not change the minimum.
\end{proof}
The count-regime floor used in \cref{sec:lt} ($\LO=10^{-30}$, $\rho\ll\LO^2$) gives, by
\eqref{eq:count-identity-main}, $k\,d_q=2n_q\log(1/\LO)+\logdet K_S+2k\log\LO$: counts again come
first, but ties are broken by the diversity of the whole subset rather than of each class.

\subsection{Tables referenced from the main text}\label{supp:sec:main-tables}
\Cref{tab:ft-dup} (full fine-tuning and near-duplicate pools), \cref{tab:coco-main} (COCO with
the COCO settings), \cref{tab:visdrone} (drone video) and \cref{tab:lvis} (LVIS) are cited in
\cref{sec:experiments-real}.

\begin{table}[!htbp]
\centering
\caption{\textbf{Full fine-tuning and near-duplicate pools}: balanced test accuracy (\%), mean{\scriptsize$\pm$}std over five seeds. Fine-tuning trains the whole CLIP image encoder on the same subsets as \cref{tab:lt-acc}; the near-duplicate pools add two lightly augmented copies of every pool image (linear probe).}
\label{tab:ft-dup}
\footnotesize
\setlength{\tabcolsep}{2.6pt}
\begin{tabular}{lcccccc}
\toprule
& \multicolumn{4}{c}{Full fine-tuning} & \multicolumn{2}{c}{Near-duplicate pool}\\
\cmidrule(lr){2-5}\cmidrule(lr){6-7}
& \multicolumn{2}{c}{CIFAR-10-LT} & \multicolumn{2}{c}{CIFAR-100-LT} & \multicolumn{2}{c}{CIFAR-100-LT}\\
\cmidrule(lr){2-3}\cmidrule(lr){4-5}\cmidrule(lr){6-7}
Selector & $k{=}200$ & $k{=}400$ & $k{=}200$ & $k{=}400$ & $k{=}200$ & $k{=}400$\\
\midrule
Uniform & 65.5{\scriptsize$\pm$4.8} & 71.5{\scriptsize$\pm$5.1} & 24.4{\scriptsize$\pm$2.0} & 34.1{\scriptsize$\pm$1.2} & 26.4{\scriptsize$\pm$1.3} & 32.7{\scriptsize$\pm$1.8}\\
Class-balanced & 88.8{\scriptsize$\pm$1.5} & 91.2{\scriptsize$\pm$0.4} & 35.3{\scriptsize$\pm$3.1} & 54.3{\scriptsize$\pm$3.5} & 42.1{\scriptsize$\pm$0.9} & 51.7{\scriptsize$\pm$1.0}\\
Ours (greedy) & 88.5{\scriptsize$\pm$1.3} & 91.0{\scriptsize$\pm$0.4} & 36.4{\scriptsize$\pm$3.0} & 54.1{\scriptsize$\pm$2.9} & 42.0{\scriptsize$\pm$0.1} & 52.1{\scriptsize$\pm$0.1}\\
\bottomrule
\end{tabular}
\end{table}

\begin{table}[!htbp]
\centering
\caption{\textbf{COCO, 80 categories} (five seeds). ``Smallest'' is the
smallest number of selected images in any category.}
\label{tab:coco-main}
\footnotesize
\setlength{\tabcolsep}{3.5pt}
\begin{tabular}{lrrrrrrrrr}
\toprule
& \multicolumn{3}{c}{$F_\tau$} & \multicolumn{3}{c}{smallest category} & \multicolumn{3}{c}{probe mAP (\%)}\\
\cmidrule(lr){2-4}\cmidrule(lr){5-7}\cmidrule(lr){8-10}
Selector & $k{=}100$ & 200 & 400 & 100 & 200 & 400 & 100 & 200 & 400\\
\midrule
Uniform & $-14.76$ & $-15.07$ & $-15.75$ & 0 & 0 & 0.3 & 33.1 & 42.3 & 51.4\\
Class-balanced & $-14.58$ & $-14.94$ & $-15.63$ & 2 & 4.8 & 10 & \textbf{38.9} & \textbf{46.9} & \textbf{52.3}\\
$k$-center & $-14.16$ & $-14.41$ & $-14.97$ & 0 & 0 & 0 & 20.4 & 30.7 & 42.4\\
Shared-kernel DPP & $-14.14$ & $-14.35$ & $-14.85$ & 0 & 0 & 0 & 19.2 & 27.9 & 43.3\\
Averaged kernel & $-14.70$ & $-14.98$ & $-15.55$ & 0 & 0 & 1 & 22.1 & 29.3 & 37.0\\
Fixed weight & $-14.65$ & $-14.89$ & $-15.31$ & 0 & 0 & 0 & 21.5 & 29.3 & 36.8\\
Ours (greedy) & $\mathbf{-13.77}$ & $\mathbf{-14.07}$ & $\mathbf{-14.64}$ & \textbf{5} & \textbf{8} & \textbf{14} & 30.6 & 39.6 & 48.6\\
\bottomrule
\end{tabular}
\end{table}

\begin{table}[!htbp]
\centering
\caption{\textbf{VisDrone, $k=20$, 15 pools.} Mean $F_\tau$; last column: pools on
which the relaxed solver beats every greedy and random-restart baseline.}
\label{tab:visdrone}
\footnotesize
\begin{tabular}{lccccc}
\toprule
$\LO$ & greedy & greedy, best $\tau'$ & random $\times200$ & relaxed + leverage & wins\\
\midrule
0.03 & $-5.614$ & $-5.597$ & $-5.608$ & $\mathbf{-5.506}$ & 15/15\\
0.05 & $-5.053$ & $-5.047$ & $-5.046$ & $\mathbf{-4.995}$ & 15/15\\
\bottomrule
\end{tabular}
\end{table}

\begin{table}[!htbp]
\centering
\caption{\textbf{LVIS, 1{,}203 categories, $k{=}400$} (protocol registered before running; five seeds). Federated average precision (\%) on rare, common and frequent categories, and the number of rare categories with at least one selected image.}
\label{tab:lvis}
\footnotesize
\setlength{\tabcolsep}{4pt}
\begin{tabular}{lrrrrr}
\toprule
Selector & AP$_r$ & AP$_c$ & AP$_f$ & AP & rare covered\\
\midrule
Uniform & 5.5 & 9.9 & 63.0 & 30.7 & 7\\
Class-balanced & 12.3 & 16.5 & 65.3 & 35.6 & 164\\
$k$-center & 5.6 & 9.4 & 56.7 & 28.0 & 12\\
Shared-kernel DPP & 5.5 & 9.4 & 56.0 & 27.7 & 12\\
Averaged kernel & 4.7 & 13.9 & 66.9 & 33.9 & 41\\
Fixed weight & 5.8 & 10.2 & 60.4 & 29.9 & 16\\
Ours (greedy) & 5.9 & 9.8 & 58.5 & 28.9 & 14\\
\bottomrule
\end{tabular}
\end{table}

\subsection{Complete results}\label{supp:sec:lt-results}
\Cref{tab:lt-tail} gives tail-class accuracy for the main comparison, \cref{tab:app-dev} the
development sweep, \cref{tab:app-coco30} COCO at the count-regime floor, and
\cref{tab:app-c10,tab:app-c100,tab:app-gtsrb,tab:app-dup100,tab:app-dup10,tab:app-c3_cifar10_r50,tab:app-c3_cifar10_r10,tab:app-c3_cifar100_r50,tab:app-c3_cifar100_r10}
every selector on every dataset and budget, with classes covered and the smallest class count.

\begin{table}[t]
\centering
\caption{\textbf{Ours against uniform and class-balanced selection, every long-tailed setting}: accuracy differences (points) and, in parentheses, seeds (of five) on which ours is higher.}
\label{tab:lt-all}
\footnotesize
\setlength{\tabcolsep}{3pt}
\begin{tabular}{lllrrrrr}
\toprule
& & & & \multicolumn{2}{c}{vs uniform} & \multicolumn{2}{c}{vs class-balanced}\\
\cmidrule(lr){5-6}\cmidrule(lr){7-8}
Data & $r$ & Role & $k$ & acc & tail & acc & tail\\
\midrule
\multirow{6}{*}{CIFAR-10-LT} & \multirow{2}{*}{100} & \multirow{2}{*}{confirm.} & 200 & \textbf{+19.7 (5)} & \textbf{+63.6 (5)} & -1.3 (0) & \textbf{+1.6 (5)}\\
 &  &  & 400 & \textbf{+19.0 (5)} & \textbf{+59.6 (5)} & -0.7 (0) & \textbf{+0.9 (5)}\\
 & \multirow{2}{*}{50} & \multirow{2}{*}{pre-reg.} & 200 & \textbf{+17.5 (5)} & \textbf{+53.6 (5)} & -0.8 (0) & \textbf{+1.6 (5)}\\
 &  &  & 400 & \textbf{+10.4 (5)} & \textbf{+35.7 (5)} & -0.8 (0) & \textbf{+1.5 (5)}\\
 & \multirow{2}{*}{10} & \multirow{2}{*}{pre-reg.} & 200 & \textbf{+3.2 (5)} & \textbf{+13.6 (5)} & -1.0 (0) & +0.7 (3)\\
 &  &  & 400 & \textbf{+1.3 (5)} & \textbf{+9.1 (5)} & -0.8 (0) & \textbf{+0.8 (4)}\\
\specialrule{0.4pt}{1pt}{1pt}
\multirow{6}{*}{CIFAR-100-LT} & \multirow{2}{*}{100} & \multirow{2}{*}{develop.} & 200 & \textbf{+15.6 (5)} & \textbf{+38.0 (5)} & -1.1 (0) & -4.7 (0)\\
 &  &  & 400 & \textbf{+19.7 (5)} & \textbf{+54.1 (5)} & -0.9 (0) & \textbf{+0.6 (4)}\\
 & \multirow{2}{*}{50} & \multirow{2}{*}{pre-reg.} & 200 & \textbf{+13.4 (5)} & \textbf{+33.2 (5)} & -1.6 (0) & -5.0 (0)\\
 &  &  & 400 & \textbf{+16.7 (5)} & \textbf{+49.2 (5)} & -1.1 (1) & +1.2 (3)\\
 & \multirow{2}{*}{10} & \multirow{2}{*}{pre-reg.} & 200 & \textbf{+9.9 (5)} & \textbf{+26.8 (5)} & \textbf{+0.2 (4)} & \textbf{+1.0 (5)}\\
 &  &  & 400 & \textbf{+7.9 (5)} & \textbf{+31.1 (5)} & -1.5 (0) & \textbf{+0.6 (4)}\\
\specialrule{0.4pt}{1pt}{1pt}
\multirow{2}{*}{GTSRB} & \multirow{2}{*}{natural} & \multirow{2}{*}{confirm.} & 200 & \textbf{+2.6 (4)} & \textbf{+28.9 (5)} & -5.2 (0) & -0.2 (2)\\
 &  &  & 400 & \textbf{+6.0 (5)} & \textbf{+35.1 (5)} & -5.1 (0) & +0.7 (3)\\
\bottomrule
\end{tabular}
\end{table}

\begin{table}[!htbp]
\centering
\caption{\textbf{Tail-class accuracy (\%)} (rarest 30\% of classes in the pool), same runs as \cref{tab:lt-acc}.}
\label{tab:lt-tail}
\footnotesize
\setlength{\tabcolsep}{2.6pt}
\begin{tabular}{lcccccc}
\toprule
& \multicolumn{2}{c}{CIFAR-10-LT ($r{=}100$)} & \multicolumn{2}{c}{CIFAR-100-LT ($r{=}100$)} & \multicolumn{2}{c}{GTSRB}\\
\cmidrule(lr){2-3}\cmidrule(lr){4-5}\cmidrule(lr){6-7}
Selector & $k{=}200$ & $k{=}400$ & $k{=}200$ & $k{=}400$ & $k{=}200$ & $k{=}400$\\
\midrule
Uniform & 31.9{\scriptsize$\pm$17.6} & 35.5{\scriptsize$\pm$18.5} & 2.1{\scriptsize$\pm$1.6} & 2.2{\scriptsize$\pm$1.7} & 14.8{\scriptsize$\pm$2.9} & 20.8{\scriptsize$\pm$3.9}\\
Class-balanced & 93.9{\scriptsize$\pm$0.7} & 94.2{\scriptsize$\pm$0.5} & \textbf{44.8{\scriptsize$\pm$1.9}} & 55.8{\scriptsize$\pm$0.7} & 43.9{\scriptsize$\pm$4.6} & 55.1{\scriptsize$\pm$2.5}\\
$k$-center & 55.7{\scriptsize$\pm$7.6} & 64.6{\scriptsize$\pm$3.4} & 6.2{\scriptsize$\pm$1.5} & 9.6{\scriptsize$\pm$1.4} & 12.2{\scriptsize$\pm$3.9} & 23.4{\scriptsize$\pm$3.5}\\
Shared-kernel DPP & 59.7{\scriptsize$\pm$0.3} & 73.5{\scriptsize$\pm$0.1} & 6.7{\scriptsize$\pm$0.0} & 6.3{\scriptsize$\pm$0.1} & 15.2{\scriptsize$\pm$0.4} & 22.1{\scriptsize$\pm$0.4}\\
Averaged kernel & 82.1{\scriptsize$\pm$0.2} & 87.3{\scriptsize$\pm$0.1} & 27.0{\scriptsize$\pm$0.2} & 37.0{\scriptsize$\pm$0.2} & \textbf{46.5{\scriptsize$\pm$0.9}} & 48.9{\scriptsize$\pm$0.6}\\
Fixed weight & 48.4{\scriptsize$\pm$0.4} & 64.7{\scriptsize$\pm$0.4} & 5.6{\scriptsize$\pm$0.0} & 6.6{\scriptsize$\pm$0.1} & 15.0{\scriptsize$\pm$0.2} & 23.9{\scriptsize$\pm$1.0}\\
Ours (greedy) & \textbf{95.5{\scriptsize$\pm$0.1}} & \textbf{95.1{\scriptsize$\pm$0.1}} & 40.1{\scriptsize$\pm$0.2} & \textbf{56.4{\scriptsize$\pm$0.2}} & 43.7{\scriptsize$\pm$0.6} & \textbf{55.9{\scriptsize$\pm$0.7}}\\
\bottomrule
\end{tabular}
\end{table}

\begin{table}[!htbp]
\centering
\caption{\textbf{Development sweep on CIFAR-100-LT, $r{=}100$} (ours only; five seeds). The first row is the registered run with the COCO settings. Accuracy and tail accuracy (\%), classes covered.}
\label{tab:app-dev}
\scriptsize
\setlength{\tabcolsep}{3pt}
\begin{tabular}{rrrrrrrr}
\toprule
& & \multicolumn{3}{c}{$k{=}200$} & \multicolumn{3}{c}{$k{=}400$}\\
\cmidrule(lr){3-5}\cmidrule(lr){6-8}
$\LO$ & $\tau$ & acc & tail & cov. & acc & tail & cov.\\
\midrule
$10^{-3}$ & 0.1 & 32.3 & 11.5 & 78 & 38.7 & 11.1 & 89\\
$10^{-3}$ & 0.03 & 29.7 & 13.7 & 92 & 40.9 & 18.1 & 100\\
$10^{-3}$ & 0.01 & 27.3 & 7.9 & 72 & 37.1 & 15.1 & 100\\
$10^{-3}$ & 0.003 & 27.7 & 8.1 & 71 & 35.2 & 9.2 & 83\\
$10^{-3}$ & 0.001 & 26.5 & 5.7 & 69 & 35.0 & 5.7 & 80\\
$10^{-6}$ & 0.1 & 38.2 & 28.5 & 100 & 48.0 & 37.3 & 100\\
$10^{-6}$ & 0.03 & 32.3 & 22.0 & 100 & 47.5 & 41.6 & 100\\
$10^{-15}$ & 0.1 & 42.5 & 39.7 & 100 & 53.3 & 54.9 & 100\\
$10^{-15}$ & 0.03 & 31.9 & 24.1 & 100 & 47.2 & 40.7 & 100\\
$10^{-30}$ & 0.1 & 42.6 & 40.1 & 100 & 53.4 & 56.4 & 100\\
$10^{-30}$ & 0.3 & 42.5 & 53.0 & 100 & 52.5 & 59.6 & 100\\
$10^{-60}$ & 0.1 & 42.6 & 40.1 & 100 & 53.1 & 57.7 & 100\\
$10^{-60}$ & 0.3 & 42.5 & 53.0 & 100 & 52.3 & 60.1 & 100\\
\bottomrule
\end{tabular}
\end{table}

\begin{table}[!htbp]
\centering
\caption{\textbf{COCO with the count-regime floor} ($\LO{=}10^{-30}$, $\tau{=}0.1$; five seeds). Probe mAP and mAP over the 20 rarest categories (\%), smallest category count, and labels per selected image. This is a separate run, so the random baselines differ slightly from \cref{tab:coco-main}.}
\label{tab:app-coco30}
\scriptsize
\setlength{\tabcolsep}{3pt}
\begin{tabular}{lrrrrrrrrr}
\toprule
& \multicolumn{3}{c}{mAP} & \multicolumn{3}{c}{mAP, rare 20} & \multicolumn{3}{c}{smallest / labels per image}\\
\cmidrule(lr){2-4}\cmidrule(lr){5-7}\cmidrule(lr){8-10}
Selector & 100 & 200 & 400 & 100 & 200 & 400 & 100 & 200 & 400\\
\midrule
Uniform & 33.1 & 42.3 & 51.6 & 21.8 & 28.0 & 37.6 & 0 / 2.9 & 0 / 2.8 & 0 / 2.9\\
Class-balanced & 38.9 & 46.9 & 52.4 & 30.3 & 39.1 & 44.8 & 2 / 3.4 & 5 / 3.4 & 10 / 3.3\\
Averaged kernel & 22.1 & 29.3 & 37.0 & 10.5 & 14.9 & 22.8 & 0 / 11.2 & 0 / 10.5 & 1 / 9.5\\
Fixed weight & 12.1 & 13.7 & 18.3 & 5.2 & 4.7 & 8.7 & 0 / 13.1 & 0 / 12.4 & 0 / 11.7\\
Ours (greedy) & 31.6 & 39.4 & 46.2 & 25.3 & 32.1 & 38.0 & 5 / 6.1 & 11 / 6.0 & 22 / 5.8\\
\bottomrule
\end{tabular}
\end{table}

\begin{table}[!htbp]
\centering
\caption{CIFAR-10-LT, $r{=}100$ (frozen settings): accuracy and tail accuracy (\%), classes covered and smallest class count.}
\label{tab:app-c10}
\scriptsize
\setlength{\tabcolsep}{2.4pt}
\begin{tabular}{lrrrrrrrrrrrr}
\toprule
& \multicolumn{4}{c}{$k{=}100$} & \multicolumn{4}{c}{$k{=}200$} & \multicolumn{4}{c}{$k{=}400$}\\
\cmidrule(lr){2-5}\cmidrule(lr){6-9}\cmidrule(lr){10-13}
Selector & acc & tail & cov. & min & acc & tail & cov. & min & acc & tail & cov. & min\\
\midrule
Uniform & 61.1{\scriptsize$\pm$7.5} & 18.2{\scriptsize$\pm$15.8} & 8 & 0.2 & 69.6{\scriptsize$\pm$5.7} & 31.9{\scriptsize$\pm$17.6} & 9 & 0.8 & 71.6{\scriptsize$\pm$6.4} & 35.5{\scriptsize$\pm$18.5} & 10 & 1.2\\
Class-balanced & 88.7{\scriptsize$\pm$0.4} & 92.6{\scriptsize$\pm$1.5} & 10 & 10.0 & 90.6{\scriptsize$\pm$0.4} & 93.9{\scriptsize$\pm$0.7} & 10 & 20.0 & 91.3{\scriptsize$\pm$0.1} & 94.2{\scriptsize$\pm$0.5} & 10 & 40.0\\
$k$-center & 67.3{\scriptsize$\pm$4.3} & 51.7{\scriptsize$\pm$6.6} & 10 & 2.2 & 73.0{\scriptsize$\pm$3.2} & 55.7{\scriptsize$\pm$7.6} & 10 & 3.2 & 78.1{\scriptsize$\pm$1.0} & 64.6{\scriptsize$\pm$3.4} & 10 & 5.0\\
Shared-kernel DPP & 64.8{\scriptsize$\pm$0.2} & 50.6{\scriptsize$\pm$0.5} & 10 & 2.0 & 73.1{\scriptsize$\pm$0.1} & 59.7{\scriptsize$\pm$0.3} & 10 & 4.0 & 83.1{\scriptsize$\pm$0.0} & 73.5{\scriptsize$\pm$0.1} & 10 & 6.0\\
Averaged kernel & 85.1{\scriptsize$\pm$0.1} & 80.4{\scriptsize$\pm$0.3} & 10 & 4.0 & 88.0{\scriptsize$\pm$0.1} & 82.1{\scriptsize$\pm$0.2} & 10 & 8.0 & 89.8{\scriptsize$\pm$0.1} & 87.3{\scriptsize$\pm$0.1} & 10 & 18.0\\
Fixed weight & 64.5{\scriptsize$\pm$0.2} & 34.2{\scriptsize$\pm$0.3} & 10 & 2.0 & 70.0{\scriptsize$\pm$0.2} & 48.4{\scriptsize$\pm$0.4} & 10 & 3.0 & 80.1{\scriptsize$\pm$0.2} & 64.7{\scriptsize$\pm$0.4} & 10 & 5.0\\
Ours (greedy) & 87.6{\scriptsize$\pm$0.1} & 96.1{\scriptsize$\pm$0.1} & 10 & 10.0 & 89.4{\scriptsize$\pm$0.1} & 95.5{\scriptsize$\pm$0.1} & 10 & 20.0 & 90.5{\scriptsize$\pm$0.0} & 95.1{\scriptsize$\pm$0.1} & 10 & 40.0\\
\bottomrule
\end{tabular}
\end{table}

\begin{table}[!htbp]
\centering
\caption{CIFAR-100-LT, $r{=}100$ (development setting), $\LO{=}10^{-30}$.}
\label{tab:app-c100}
\scriptsize
\setlength{\tabcolsep}{2.4pt}
\begin{tabular}{lrrrrrrrr}
\toprule
& \multicolumn{4}{c}{$k{=}200$} & \multicolumn{4}{c}{$k{=}400$}\\
\cmidrule(lr){2-5}\cmidrule(lr){6-9}
Selector & acc & tail & cov. & min & acc & tail & cov. & min\\
\midrule
Uniform & 27.0{\scriptsize$\pm$0.9} & 2.1{\scriptsize$\pm$1.6} & 61 & 0.0 & 33.7{\scriptsize$\pm$1.2} & 2.2{\scriptsize$\pm$1.7} & 70 & 0.0\\
Class-balanced & 43.7{\scriptsize$\pm$0.7} & 44.8{\scriptsize$\pm$1.9} & 100 & 2.0 & 54.2{\scriptsize$\pm$0.9} & 55.8{\scriptsize$\pm$0.7} & 100 & 4.0\\
$k$-center & 28.4{\scriptsize$\pm$0.9} & 6.2{\scriptsize$\pm$1.5} & 66 & 0.0 & 35.6{\scriptsize$\pm$1.0} & 9.6{\scriptsize$\pm$1.4} & 79 & 0.0\\
Shared-kernel DPP & 27.3{\scriptsize$\pm$0.0} & 6.7{\scriptsize$\pm$0.0} & 67 & 0.0 & 33.6{\scriptsize$\pm$0.1} & 6.3{\scriptsize$\pm$0.1} & 74 & 0.0\\
Averaged kernel & 39.7{\scriptsize$\pm$0.1} & 27.0{\scriptsize$\pm$0.2} & 100 & 1.0 & 50.8{\scriptsize$\pm$0.1} & 37.0{\scriptsize$\pm$0.2} & 100 & 1.0\\
Fixed weight & 27.0{\scriptsize$\pm$0.1} & 5.6{\scriptsize$\pm$0.0} & 68 & 0.0 & 34.1{\scriptsize$\pm$0.2} & 6.6{\scriptsize$\pm$0.1} & 77 & 0.0\\
Ours (greedy) & 42.6{\scriptsize$\pm$0.1} & 40.1{\scriptsize$\pm$0.2} & 100 & 1.0 & 53.4{\scriptsize$\pm$0.2} & 56.4{\scriptsize$\pm$0.2} & 100 & 3.0\\
\bottomrule
\end{tabular}
\end{table}

\begin{table}[!htbp]
\centering
\caption{GTSRB, natural imbalance (frozen settings).}
\label{tab:app-gtsrb}
\scriptsize
\setlength{\tabcolsep}{2.4pt}
\begin{tabular}{lrrrrrrrrrrrr}
\toprule
& \multicolumn{4}{c}{$k{=}100$} & \multicolumn{4}{c}{$k{=}200$} & \multicolumn{4}{c}{$k{=}400$}\\
\cmidrule(lr){2-5}\cmidrule(lr){6-9}\cmidrule(lr){10-13}
Selector & acc & tail & cov. & min & acc & tail & cov. & min & acc & tail & cov. & min\\
\midrule
Uniform & 31.7{\scriptsize$\pm$1.3} & 7.4{\scriptsize$\pm$2.8} & 34 & 0.0 & 42.0{\scriptsize$\pm$2.0} & 14.8{\scriptsize$\pm$2.9} & 40 & 0.0 & 50.3{\scriptsize$\pm$2.7} & 20.8{\scriptsize$\pm$3.9} & 41 & 0.2\\
Class-balanced & 40.7{\scriptsize$\pm$1.9} & 38.2{\scriptsize$\pm$6.2} & 43 & 2.0 & 49.8{\scriptsize$\pm$2.1} & 43.9{\scriptsize$\pm$4.6} & 43 & 4.0 & 61.5{\scriptsize$\pm$1.5} & 55.1{\scriptsize$\pm$2.5} & 43 & 9.0\\
$k$-center & 30.7{\scriptsize$\pm$1.9} & 8.7{\scriptsize$\pm$2.3} & 35 & 0.0 & 39.1{\scriptsize$\pm$1.8} & 12.2{\scriptsize$\pm$3.9} & 41 & 0.0 & 50.8{\scriptsize$\pm$1.4} & 23.4{\scriptsize$\pm$3.5} & 43 & 0.6\\
Shared-kernel DPP & 30.2{\scriptsize$\pm$0.3} & 5.4{\scriptsize$\pm$0.6} & 36 & 0.0 & 41.4{\scriptsize$\pm$0.1} & 15.2{\scriptsize$\pm$0.4} & 40 & 0.0 & 49.5{\scriptsize$\pm$0.2} & 22.1{\scriptsize$\pm$0.4} & 43 & 1.0\\
Averaged kernel & 40.3{\scriptsize$\pm$0.1} & 28.8{\scriptsize$\pm$0.5} & 43 & 1.0 & 54.8{\scriptsize$\pm$0.3} & 46.5{\scriptsize$\pm$0.9} & 43 & 2.0 & 62.3{\scriptsize$\pm$0.2} & 48.9{\scriptsize$\pm$0.6} & 43 & 5.0\\
Fixed weight & 30.6{\scriptsize$\pm$0.2} & 8.1{\scriptsize$\pm$0.5} & 36 & 0.0 & 42.1{\scriptsize$\pm$0.1} & 15.0{\scriptsize$\pm$0.2} & 40 & 0.0 & 49.3{\scriptsize$\pm$0.5} & 23.9{\scriptsize$\pm$1.0} & 43 & 1.0\\
Ours (greedy) & 35.7{\scriptsize$\pm$0.4} & 33.6{\scriptsize$\pm$0.5} & 43 & 2.0 & 44.6{\scriptsize$\pm$0.1} & 43.7{\scriptsize$\pm$0.6} & 43 & 4.0 & 56.3{\scriptsize$\pm$0.3} & 55.9{\scriptsize$\pm$0.7} & 43 & 9.0\\
\bottomrule
\end{tabular}
\end{table}

\begin{table}[!htbp]
\centering
\caption{CIFAR-100-LT, $r{=}100$, with two near-duplicate copies of every pool image.}
\label{tab:app-dup100}
\scriptsize
\setlength{\tabcolsep}{2.4pt}
\begin{tabular}{lrrrrrrrr}
\toprule
& \multicolumn{4}{c}{$k{=}200$} & \multicolumn{4}{c}{$k{=}400$}\\
\cmidrule(lr){2-5}\cmidrule(lr){6-9}
Selector & acc & tail & cov. & min & acc & tail & cov. & min\\
\midrule
Uniform & 26.4{\scriptsize$\pm$1.3} & 2.1{\scriptsize$\pm$0.9} & 61 & 0.0 & 32.7{\scriptsize$\pm$1.8} & 3.0{\scriptsize$\pm$2.4} & 71 & 0.0\\
Class-balanced & 42.1{\scriptsize$\pm$0.9} & 41.5{\scriptsize$\pm$0.8} & 100 & 2.0 & 51.7{\scriptsize$\pm$1.0} & 50.8{\scriptsize$\pm$1.1} & 100 & 4.0\\
$k$-center & 26.9{\scriptsize$\pm$1.2} & 5.7{\scriptsize$\pm$1.6} & 65 & 0.0 & 34.9{\scriptsize$\pm$0.8} & 8.1{\scriptsize$\pm$1.9} & 79 & 0.0\\
Shared-kernel DPP & 24.8{\scriptsize$\pm$0.0} & 6.0{\scriptsize$\pm$0.1} & 65 & 0.0 & 33.0{\scriptsize$\pm$0.1} & 7.8{\scriptsize$\pm$0.2} & 75 & 0.0\\
Averaged kernel & 40.2{\scriptsize$\pm$0.1} & 27.3{\scriptsize$\pm$0.2} & 100 & 1.0 & 49.8{\scriptsize$\pm$0.1} & 35.6{\scriptsize$\pm$0.2} & 100 & 1.0\\
Fixed weight & 24.0{\scriptsize$\pm$0.1} & 6.6{\scriptsize$\pm$0.2} & 66 & 0.0 & 32.7{\scriptsize$\pm$0.1} & 7.6{\scriptsize$\pm$0.1} & 79 & 0.0\\
Ours (greedy) & 42.0{\scriptsize$\pm$0.1} & 44.7{\scriptsize$\pm$0.2} & 100 & 1.0 & 52.1{\scriptsize$\pm$0.1} & 56.6{\scriptsize$\pm$0.1} & 100 & 3.0\\
\bottomrule
\end{tabular}
\end{table}

\begin{table}[!htbp]
\centering
\caption{CIFAR-10-LT, $r{=}100$, with two near-duplicate copies of every pool image.}
\label{tab:app-dup10}
\scriptsize
\setlength{\tabcolsep}{2.4pt}
\begin{tabular}{lrrrrrrrr}
\toprule
& \multicolumn{4}{c}{$k{=}200$} & \multicolumn{4}{c}{$k{=}400$}\\
\cmidrule(lr){2-5}\cmidrule(lr){6-9}
Selector & acc & tail & cov. & min & acc & tail & cov. & min\\
\midrule
Uniform & 66.3{\scriptsize$\pm$4.1} & 21.8{\scriptsize$\pm$11.0} & 9 & 0.2 & 72.9{\scriptsize$\pm$5.5} & 37.6{\scriptsize$\pm$15.5} & 10 & 0.8\\
Class-balanced & 90.0{\scriptsize$\pm$0.5} & 94.4{\scriptsize$\pm$0.8} & 10 & 20.0 & 91.1{\scriptsize$\pm$0.4} & 95.1{\scriptsize$\pm$0.9} & 10 & 40.0\\
$k$-center & 72.7{\scriptsize$\pm$2.7} & 60.0{\scriptsize$\pm$11.4} & 10 & 3.8 & 79.2{\scriptsize$\pm$0.3} & 70.9{\scriptsize$\pm$1.1} & 10 & 6.6\\
Shared-kernel DPP & 73.8{\scriptsize$\pm$0.1} & 68.3{\scriptsize$\pm$0.3} & 10 & 4.0 & 82.9{\scriptsize$\pm$0.1} & 76.3{\scriptsize$\pm$0.2} & 10 & 7.0\\
Averaged kernel & 87.2{\scriptsize$\pm$0.1} & 87.3{\scriptsize$\pm$0.1} & 10 & 10.0 & 90.2{\scriptsize$\pm$0.1} & 90.4{\scriptsize$\pm$0.1} & 10 & 19.0\\
Fixed weight & 71.5{\scriptsize$\pm$0.1} & 56.3{\scriptsize$\pm$0.2} & 10 & 3.0 & 80.1{\scriptsize$\pm$0.1} & 70.7{\scriptsize$\pm$0.2} & 10 & 5.0\\
Ours (greedy) & 89.5{\scriptsize$\pm$0.1} & 95.5{\scriptsize$\pm$0.0} & 10 & 20.0 & 90.6{\scriptsize$\pm$0.1} & 95.9{\scriptsize$\pm$0.1} & 10 & 40.0\\
\bottomrule
\end{tabular}
\end{table}

\begin{table}[!htbp]
\centering
\caption{CIFAR-10-LT, $r{=}50$ (pre-registered tail test; three selectors).}
\label{tab:app-c3_cifar10_r50}
\scriptsize
\setlength{\tabcolsep}{2.4pt}
\begin{tabular}{lrrrrrrrr}
\toprule
& \multicolumn{4}{c}{$k{=}200$} & \multicolumn{4}{c}{$k{=}400$}\\
\cmidrule(lr){2-5}\cmidrule(lr){6-9}
Selector & acc & tail & cov. & min & acc & tail & cov. & min\\
\midrule
Uniform & 72.0{\scriptsize$\pm$5.7} & 42.0{\scriptsize$\pm$13.2} & 9 & 0.8 & 80.3{\scriptsize$\pm$5.0} & 60.2{\scriptsize$\pm$15.3} & 10 & 2.6\\
Class-balanced & 90.3{\scriptsize$\pm$0.5} & 94.0{\scriptsize$\pm$0.6} & 10 & 20.0 & 91.5{\scriptsize$\pm$0.2} & 94.4{\scriptsize$\pm$0.5} & 10 & 40.0\\
Ours (greedy) & 89.5{\scriptsize$\pm$0.0} & 95.7{\scriptsize$\pm$0.1} & 10 & 20.0 & 90.7{\scriptsize$\pm$0.1} & 95.9{\scriptsize$\pm$0.0} & 10 & 40.0\\
\bottomrule
\end{tabular}
\end{table}

\begin{table}[!htbp]
\centering
\caption{CIFAR-10-LT, $r{=}10$ (pre-registered tail test; three selectors).}
\label{tab:app-c3_cifar10_r10}
\scriptsize
\setlength{\tabcolsep}{2.4pt}
\begin{tabular}{lrrrrrrrr}
\toprule
& \multicolumn{4}{c}{$k{=}200$} & \multicolumn{4}{c}{$k{=}400$}\\
\cmidrule(lr){2-5}\cmidrule(lr){6-9}
Selector & acc & tail & cov. & min & acc & tail & cov. & min\\
\midrule
Uniform & 86.2{\scriptsize$\pm$1.1} & 81.3{\scriptsize$\pm$4.1} & 10 & 4.2 & 89.2{\scriptsize$\pm$0.7} & 86.7{\scriptsize$\pm$1.6} & 10 & 9.4\\
Class-balanced & 90.4{\scriptsize$\pm$0.6} & 94.3{\scriptsize$\pm$0.9} & 10 & 20.0 & 91.3{\scriptsize$\pm$0.4} & 94.9{\scriptsize$\pm$0.8} & 10 & 40.0\\
Ours (greedy) & 89.4{\scriptsize$\pm$0.0} & 95.0{\scriptsize$\pm$0.1} & 10 & 20.0 & 90.5{\scriptsize$\pm$0.1} & 95.7{\scriptsize$\pm$0.0} & 10 & 40.0\\
\bottomrule
\end{tabular}
\end{table}

\begin{table}[!htbp]
\centering
\caption{CIFAR-100-LT, $r{=}50$ (pre-registered tail test; three selectors).}
\label{tab:app-c3_cifar100_r50}
\scriptsize
\setlength{\tabcolsep}{2.4pt}
\begin{tabular}{lrrrrrrrr}
\toprule
& \multicolumn{4}{c}{$k{=}200$} & \multicolumn{4}{c}{$k{=}400$}\\
\cmidrule(lr){2-5}\cmidrule(lr){6-9}
Selector & acc & tail & cov. & min & acc & tail & cov. & min\\
\midrule
Uniform & 28.4{\scriptsize$\pm$1.5} & 4.8{\scriptsize$\pm$2.3} & 65 & 0.0 & 36.0{\scriptsize$\pm$0.8} & 5.9{\scriptsize$\pm$1.9} & 79 & 0.0\\
Class-balanced & 43.5{\scriptsize$\pm$0.3} & 43.0{\scriptsize$\pm$2.3} & 100 & 2.0 & 53.8{\scriptsize$\pm$0.9} & 53.8{\scriptsize$\pm$1.3} & 100 & 4.0\\
Ours (greedy) & 41.8{\scriptsize$\pm$0.2} & 38.0{\scriptsize$\pm$0.2} & 100 & 1.0 & 52.7{\scriptsize$\pm$0.1} & 55.0{\scriptsize$\pm$0.3} & 100 & 3.0\\
\bottomrule
\end{tabular}
\end{table}

\begin{table}[!htbp]
\centering
\caption{CIFAR-100-LT, $r{=}10$ (pre-registered tail test; three selectors).}
\label{tab:app-c3_cifar100_r10}
\scriptsize
\setlength{\tabcolsep}{2.4pt}
\begin{tabular}{lrrrrrrrr}
\toprule
& \multicolumn{4}{c}{$k{=}200$} & \multicolumn{4}{c}{$k{=}400$}\\
\cmidrule(lr){2-5}\cmidrule(lr){6-9}
Selector & acc & tail & cov. & min & acc & tail & cov. & min\\
\midrule
Uniform & 32.8{\scriptsize$\pm$2.3} & 14.0{\scriptsize$\pm$2.7} & 78 & 0.0 & 44.2{\scriptsize$\pm$0.9} & 21.8{\scriptsize$\pm$3.3} & 92 & 0.0\\
Class-balanced & 42.6{\scriptsize$\pm$0.4} & 39.7{\scriptsize$\pm$0.7} & 100 & 2.0 & 53.6{\scriptsize$\pm$0.2} & 52.3{\scriptsize$\pm$2.2} & 100 & 4.0\\
Ours (greedy) & 42.8{\scriptsize$\pm$0.1} & 40.8{\scriptsize$\pm$0.3} & 100 & 1.0 & 52.1{\scriptsize$\pm$0.1} & 52.9{\scriptsize$\pm$0.2} & 100 & 3.0\\
\bottomrule
\end{tabular}
\end{table}

\subsection{Additional baselines and the zero-padding endpoint}\label{supp:sec:extra-baselines}
\paragraph{Baselines.} The quota-constrained \DPP, in the spirit of \citet{celis2018fair} and
\citet{nikolov2016partition}, fixes the class-balanced allocation (equal counts up to one, a class
contributing all its members if it has fewer) and, within it, greedily maximizes $\logdet K_S$.
SATURATE \citep{krause2008robust} maximizes $\min_qF_q(S)$ for the monotone submodular
$F_q(S)=\logdet(I+K_{S_q})$: a binary search over a level $c$, each step running greedy on
$\sum_q\min(F_q,c)$; unused budget is filled greedily. Pools, test sets, probe, seeds and $k$ are
those of \cref{tab:lt-all}. These selectors, like ours, are deterministic, so across seeds only the
probe initialization changes.

\paragraph{Results.} With 20--40 items per class (CIFAR-10-LT), ours is more accurate on tail classes
than the quota \DPP\ in all six settings and more accurate overall than SATURATE in five
(\cref{supp:tab:extra}). With 2--10 items per class (CIFAR-100-LT, GTSRB), the quota \DPP\ is more
accurate on tail classes and SATURATE overall. Both diversify within each class, whereas our count
regime breaks ties by the diversity of the whole subset (\cref{supp:sec:zero-cb}), which matters
little with many items per class and much with few. The quota \DPP\ also needs disjoint groups and
counts fixed in advance; with overlapping categories one item fills several quotas, whereas our
objective and SATURATE apply directly.

\paragraph{Zero-padding endpoint (exploratory).} After seeing these results we ran our objective at
the zero-padding endpoint of \cref{prop:zero-cb}, which diversifies within each class: exact scores
$\logdet(K_{S_q}+\rho I)+(k-n_q)\log\rho$, greedy, $\tau=0.1$, and $\rho=10^{-60}$, chosen so that one
extra member changes a category score by $\log(1/\rho)=138$, the same step as $2\log(1/\LO)$ in the
count regime; no other $\rho$ was run. It improves ours on CIFAR-100-LT (overall in five of six
settings, tail in all six) and on GTSRB, by about 8 points overall and 11--12 on tail classes, where
it is within 0.7 points of SATURATE overall, more accurate on tail classes, and 3 points (overall) and
10--13 points (tail) above class-balanced sampling. On CIFAR-10-LT the count regime remains better in most settings.
The floor thus selects between two regimes of the same objective: the count regime when classes
receive many items, zero padding when they receive few.

\begin{table}[!htbp]
\centering
\caption{\textbf{Additional baselines} (five seeds): balanced accuracy / tail accuracy (\%). ``Zero
padding'' is our objective at $\LO=0$, $\rho=10^{-60}$ (exploratory, chosen after the baselines).}
\label{supp:tab:extra}
\scriptsize
\setlength{\tabcolsep}{3pt}
\begin{tabular}{llcccccc}
\toprule
Data & $k$ & Ours & Class-balanced & Quota \DPP & SATURATE & Zero padding\\
\midrule
CIFAR-10-LT, $r{=}100$ & 200 & 89.4 / 95.5 & 90.6 / 93.9 & 89.6 / 94.1 & 88.8 / 94.8 & 88.0 / 93.7\\
 & 400 & 90.5 / 95.1 & 91.3 / 94.2 & 91.3 / 94.2 & 89.3 / 94.1 & 90.0 / 93.8\\
CIFAR-10-LT, $r{=}50$ & 200 & 89.5 / 95.7 & 90.3 / 94.0 & 89.3 / 93.8 & 90.2 / 94.3 & 89.5 / 94.4\\
 & 400 & 90.7 / 95.9 & 91.5 / 94.4 & 91.0 / 95.3 & 89.2 / 95.3 & 90.0 / 94.6\\
CIFAR-10-LT, $r{=}10$ & 200 & 89.4 / 95.0 & 90.4 / 94.3 & 89.8 / 94.7 & 88.8 / 96.6 & 89.5 / 96.6\\
 & 400 & 90.5 / 95.7 & 91.3 / 94.9 & 90.9 / 95.6 & 90.1 / 96.8 & 90.4 / 96.3\\
CIFAR-100-LT, $r{=}100$ & 200 & 42.6 / 40.1 & 43.7 / 44.8 & 44.3 / 54.9 & 43.0 / 41.5 & 43.8 / 45.1\\
 & 400 & 53.4 / 56.4 & 54.2 / 55.8 & 52.6 / 61.8 & 53.6 / 55.4 & 54.3 / 58.6\\
CIFAR-100-LT, $r{=}50$ & 200 & 41.8 / 38.0 & 43.5 / 43.0 & 43.1 / 52.4 & 43.8 / 45.6 & 44.1 / 44.7\\
 & 400 & 52.7 / 55.0 & 53.8 / 53.8 & 52.2 / 58.7 & 54.5 / 55.7 & 53.7 / 56.1\\
CIFAR-100-LT, $r{=}10$ & 200 & 42.8 / 40.8 & 42.6 / 39.7 & 41.2 / 42.1 & 43.8 / 43.5 & 42.2 / 42.1\\
 & 400 & 52.1 / 52.9 & 53.6 / 52.3 & 53.4 / 54.9 & 53.5 / 53.5 & 53.6 / 56.8\\
GTSRB (natural) & 200 & 44.6 / 43.7 & 49.8 / 43.9 & 47.3 / 45.6 & 53.2 / 52.0 & 52.9 / 54.5\\
 & 400 & 56.3 / 55.9 & 61.5 / 55.1 & 58.2 / 58.2 & 64.9 / 65.5 & 64.2 / 67.8\\
\bottomrule
\end{tabular}
\end{table}

\section{Independent Selection and Kernel Calibration}\label{supp:sec:model}\label{supp:sec:group-selection}

\subsection{Overlapping categories and independent selection}\label{supp:sec:independent-selection}
Consider a dataset indexed by $[n]$ with $M$ nonempty categories
$G_1,\ldots,G_M\subseteq[n]$. We use \emph{category} and \emph{group}
interchangeably. Memberships may overlap: the same item can belong to several
categories. Given a subset size $k$, our goal is to retain $k$ distinct items that
provide adequate representation across the categories while preserving
feature diversity. The categories may differ substantially in size,
redundancy, and overlap, making the allocation of these $k$ places a
central part of the selection problem.

As a running example, consider images labeled by the objects they show
(\cref{fig:intro-example}). An image containing a dog and a car belongs to
both the dog and car categories. Selecting it occupies one of the $k$ places
but contributes an example to each category. The same situation arises in any dataset whose items can carry several labels. For a selected set $S$, write
$n_q(S):=|S\cap G_q|$. These counts measure category representation; their
sum can exceed $|S|$ because a single item may contribute to several groups.

A natural starting point is to run an independent \DPP-based selector on
the items belonging to each category. Let $u_i$ denote a shared unit feature
vector for item $i$, let $K_{ij}:=\langle u_i,u_j\rangle$, and write
$K^q:=K_{G_q}$ for the kernel restricted to category $q$. The same feature
representation is used for all categories; category membership does not
require a separate modality or feature space. Prescribing a size $k_q$
for each category gives the following baseline.

\begin{definition}[Independent category-wise selection]\label{supp:def:stratified}
For each category $q$, choose $1\le k_q\le |G_q|$ and solve its fixed-size
\DPP\ MAP problem independently (assuming $\rank(K_{G_q})\ge k_q$):
\[
  S_q^\star\in\operatorname*{arg\,max}_{S_q\subseteq G_q,\ |S_q|=k_q}
  \logdet(K_{S_q}).
\]
The resulting selection is $S_{\mathrm{ind}}=\bigcup_{q=1}^M S_q^\star$.
\end{definition}

In the running example, one selector chooses diverse dog images and
another chooses diverse car images. This approach addresses diversity
within each category, but leaves three difficulties.

First, \emph{overlapping memberships can produce repeated selections}.
Both selectors may choose the same dog-and-car image. That item
represents both categories but remains a single selected example, so even
when $\sum_q k_q=k$, the union can contain fewer than $k$ distinct items.
Removing duplicates therefore leaves a further allocation decision about
how to fill the remaining places.

Second, \emph{the category sizes must be fixed in advance}. The choice
of $k_q$ depends on category sizes, internal diversity, and overlap with
other categories. Equal sizes need not reflect these differences, while
unequal sizes require an allocation rule before the selected examples
are known. In particular, deciding the numbers of dog and car images
separately does not account for which examples can represent both.

Third, \emph{the selectors do not coordinate their decisions}. A dog
selector evaluates its candidates without accounting for what the car
selector retains or which other categories remain poorly represented.
Consequently, an item's contribution to several categories and its
redundancy with examples selected elsewhere are not assessed together.

\begin{remark}[Kernel calibration]\label{supp:rem:calibration}
Categories can only be compared on a common scale. All kernels here use the
same $K$, $\LO$, and $\rho$, fixed before selection. Multiplying one category's
kernel by $c>0$ adds $\log c$ to its category diversity and can change which
category is least diverse, so the kernels must not be rescaled separately.
\end{remark}

\section{Locally Optimal Conjugate Gradient Acceleration}
\label{supp:sec:locg}

The COCO solver study leaves open whether slow convergence of the relaxed
problem accounts for its poorer discrete subsets. To test this, we apply locally
optimal conjugate gradient (LOCG) acceleration to the nonlinear SCF iteration
\citep{li2024nepv,wang2026nepvadmm}.
LOCG reuses the preceding frame and takes several nonlinear SCF steps
in a small search space before rebuilding it.

\paragraph{Restricted problem.}
Let $W_t$ have orthonormal columns spanning
$[V_t, A_\perp(V_t), V_{t-1}]$, omitting $V_{t-1}$ on the first iteration
and taking $V_t$ as the first $k$ columns. Write $V=W_tY$, where
$Y\in\Stiefel(m,k)$ and $m\le 3k$. For each category, form
$B_{q,t}=W_t\T L^{(q)}W_t$ once, using weighted features:
\begin{equation}
B_{q,t}=(U\T\Omega_q W_t)\T(U\T\Omega_q W_t)+\rho I_m.
\end{equation}
The restricted category score is $k^{-1}\logdet(Y\T B_{q,t}Y)$, exactly
$d_q(W_tY)$. Each inner SCF sweep recomputes these scores, their soft-min
weights and the projected residual
$R_t(Y)=W_t\T A_\perp(W_tY)$. It takes the leading $k$ eigenvectors of
$(1/k+\sigma)YY\T + R_t(Y)Y\T + Y R_t(Y)\T$, then applies the same
alignment and damping as \cref{alg:scf}. We start with the first $k$ columns
of $I_m$, take five sweeps, and set $V_{t+1}=W_tY$. The full-space residual
$\|A_\perp(V)\|_F$ determines stopping. The shift and damping are identical
to the corresponding SCF baseline. We use the fixed inner budget without
an additional objective acceptance test.

\paragraph{Why the acceleration is nonlinear.}
If the linear lift is frozen at $V_t$, its initial block Ritz residual
from $V_t$ is $A_\perp(V_t)$, since $V_t\T H_{\tau,\sigma}V_t=(1/k+\sigma)I_k$.
The span of the starting block and this residual contains the entire range of
$H_{\tau,\sigma}$. Consequently, one unpreconditioned block Rayleigh--Ritz
expansion of a warm-started LOBPCG eigensolver \citep{knyazev2001lobpcg} already
recovers the exact dominant subspace in exact arithmetic. This is the same
eigenspace computed by the $2k$ compression in \cref{prop:gap}. LOCG changes
the nonlinear iteration by updating the operator and its category weights
within the search space. With one inner sweep, it gives the same update as
SCF in exact arithmetic when the search space contains the residual.

\paragraph{Paired GPU protocol.}
Both methods use the same double-precision PyTorch backend on one GeForce
RTX 5080, the same precomputed CLIP features, initial frames, rounding rules,
and dataset-specific v3 stopping configurations. The COCO pilot uses the
one-eighth scaling pool ($13{,}408$ candidates), $k=100$, $\tau=0.1$,
$\LO=10^{-3}$, and the original $10{,}000$-image test split. It uses
a $4k$ leverage shortlist, greedy completion and swaps, followed by the
same multi-label probe as \cref{supp:sec:coco-setup}. The VisDrone pilot
uses three of the existing pools (seeds 5--7), $k=20$, $\tau=0.05$,
and both $\LO=0.03$ and $0.05$. The budget check uses seeds 8--12 at
$k=15,20,25$ and $\LO=0.05$. Both relaxations use top-$k$ leverage
rounding and identical one-swap refinement. VisDrone measures the discrete
objective. Selection times exclude feature extraction and probe fitting;
greedy and rounding use the host CPU. We check agreement with the original
SCF update and verify that rotating the reduced basis leaves the restricted
scores and lifted subspace unchanged to numerical precision.

\subsection{A sufficient level shift for ascent}\label{sec:shift-proof}

Fix $k$, $\rho>0$ and $\tau>0$. For a symmetric matrix $P$ with
$0\preceq P\preceq I$, extend the category score by
\[
\bar d_q(P)=k^{-1}\log\det\bigl(I+(L^{(q)}-I)P\bigr),
\qquad g(P)=\SoftMin_\tau\bigl((\bar d_q(P))_{q=1}^M\bigr).
\]
At $P=VV\T$, writing the matrix inside the determinant in the basis
$[V,V_\perp]$ gives a block triangular matrix with diagonal blocks
$V\T L^{(q)}V$ and $I$. Hence $g(VV\T)=f_\tau(V)$.
Sylvester's determinant identity gives the equivalent determinant of
$I-P+P^{1/2}L^{(q)}P^{1/2}$. This matrix is positive definite throughout
$0\preceq P\preceq I$, since $L^{(q)}\succeq \rho I$. The determinant is
therefore positive and $g$ is smooth on a neighborhood of the compact
convex set. Let $B$ bound $\|\nabla g\|_2$ there and $C$ bound the
Hessian operator norm for the Frobenius metric.

\begin{proposition}[Level shift implies the NEPv Ansatz]\label{prop:shift-ansatz}
Let $P=VV\T$ and $Q$ be rank-$k$ orthogonal projectors, set
$\Delta=Q-P$, and use the lift $H_{\tau,\sigma}(V)$ of \cref{prop:gap}.
Then
\[
g(Q)-g(P)\ \ge\ \tr\bigl(H_{\tau,\sigma}(V)\Delta\bigr)
             + \frac{\beta-2B-C}{2}\|\Delta\|_F^2 .
\]
In particular, for $\beta\ge 2B+C$ the frozen trace gain certifies at
least as much objective gain. This gives Li's NEPv Ansatz with
$\omega=1$ for the full update and every exact restricted subproblem.
\end{proposition}
\begin{proof}
Write $G=\nabla g(P)$. Differentiating $f_\tau(V)=g(VV\T)$ along
horizontal frame tangents shows that $(I-P)GV=A_\perp(V)$.
Thus $G-H_{\tau,\sigma}$ is block diagonal in the decomposition
$\operatorname{range}(P)\oplus\ker(P)$. In that decomposition,
$P\Delta P\preceq0$ and $(I-P)\Delta(I-P)\succeq0$, with equal
absolute traces $a=k-\tr(PQ)=\|\Delta\|_F^2/2$. The diagonal
blocks of $H_{\tau,\sigma}$ are $\beta I$ and zero.
Using $\|G\|_2\le B$ gives
$\tr((G-H_{\tau,\sigma})\Delta)\ge (\beta-2B)a$.
Taylor's formula along the segment from $P$ to $Q$ bounds the remainder
below by $-C\|\Delta\|_F^2/2$, proving the inequality. It applies to
projectors in any fixed restricted subspace by substituting
$P=WYY\T W\T$ and $Q=W\widehat Y\widehat Y\T W\T$.
\end{proof}

The top-eigenspace update increases the frozen trace by the variational
principle. The aligned, polar-damped update also increases it. In each
principal plane of \cref{prop:gap}, with residual singular value $s$,
the trace gain at angle $\phi$ equals
$2s\sin\phi\cos\phi-\beta\sin^2\phi$. This function is nondecreasing
from zero up to its maximizer $\theta=\tfrac12\arctan(2s/\beta)$.
The damped angle is
$\arctan\{\alpha\sin\theta/(1-\alpha+\alpha\cos\theta)\}\in[0,\theta]$.
The same reasoning applies to each inner restricted sweep. The condition
is sufficient and may be conservative. It supplies neither a numerical
convergence rate at the tested shifts nor an assurance about rounding.

\paragraph{A conservative bound.}
Let $a=\max_q\|L^{(q)}-I\|_2$, $m=\min(1,\rho)$, and
$r=a+a^2/m$. For $A=L^{(q)}-I$, the matrix
$T_q=(I+AP)^{-1}A$ is symmetric. The inverse identity
$T_q=A-AP^{1/2}(I+P^{1/2}AP^{1/2})^{-1}P^{1/2}A$ gives
$\|T_q\|_2\le r$. The category gradient is $T_q/k$, and its
Hessian bilinear form is $-k^{-1}\tr(T_qE T_qF)$. The soft-min
gradient is a convex combination of the category gradients.
Its Hessian adds their negative covariance divided by $\tau$ to
the weighted average of their Hessians. Thus valid bounds are
$B=r/k$ and $C=r^2/k+nr^2/(\tau k^2)$. The covariance bound here
uses $\|T_q/k\|_F\le\sqrt n\,r/k$.

\paragraph{Local rate and the spectral gap.}
At a stationary subspace, the principal angles give
$\log_V(V^+)=(\alpha/\beta)A_\perp+o(\|A_\perp\|_F)$, where
$\log_V$ is the Grassmann logarithm. Using
$\operatorname{grad}f_\tau=2A_\perp$, the differential of the damped SCF
map at that subspace is $I+\alpha\,\operatorname{Hess}f_\tau/(2\beta)$.
At a nondegenerate local maximum with Hessian eigenvalues $-\mu_i<0$,
the linearized factors are $1-\alpha\mu_i/(2\beta)$. Their magnitudes
are below one if $\beta>\alpha\max_i\mu_i/4$. For very large shifts the
factors approach one even as the frozen gap increases. At a degenerate
stationary point this calculation can give a unit factor. This dependence
on the objective curvature is separate from the frozen gap
\citep{cai2018eigenvector,baililu2022scf}.

\subsection{Convergence traces and additional comparisons}
\begin{figure}[!htbp]
\centering
\includegraphics[width=0.98\linewidth]{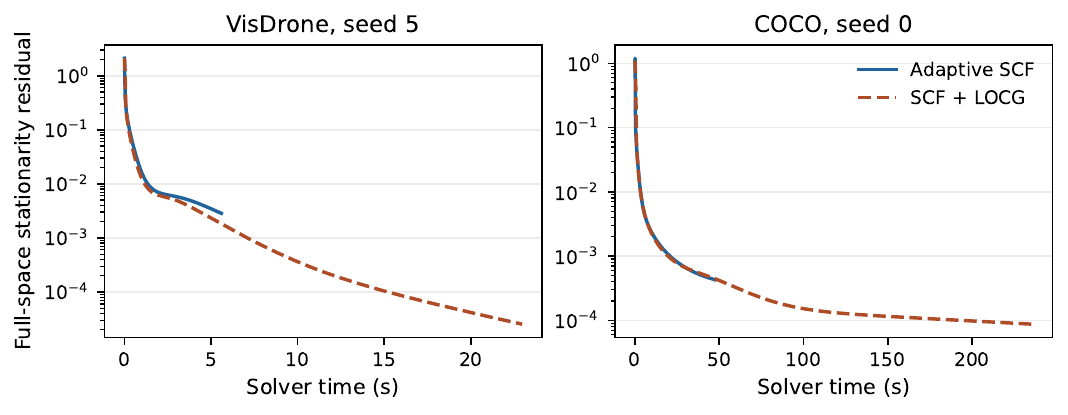}
\caption{\textbf{Full-space residual versus solver time.}
Representative matched runs from \cref{tab:locg-main}. VisDrone uses floor
$0.05$. Timings exclude rounding and probe training.}
\label{fig:locg-residual}
\end{figure}
For the primary VisDrone results and the COCO pilot, all endpoints reach
the iteration cap. Per run, adaptive SCF uses 500 full updates on VisDrone
and 300 on COCO. LOCG uses the same number of outer updates and reduced
kernel builds, with five projected SCF sweeps per outer update. Solver
timing includes the full-space function and residual evaluations. On COCO
the reduced solves account for $166.4$ seconds and search-space construction
and kernel compression account for $59.4$ seconds of LOCG solver time.
The full-space evaluations take $10.1$ seconds for either solver. The
difference between residual and selection outcomes is visible in
\cref{fig:locg-residual,tab:locg-main}.

\begin{table}[!htbp]
\centering
\caption{\textbf{Level-shift diagnostic on the paired COCO pool.}
Each run takes 100 damped SCF updates from the same frame. We count
decreases greater than $10^{-9}$ in magnitude. Solve time excludes
eigengap diagnostics and rounding.}
\label{tab:locg-shifts}
\small
\begin{tabular}{rrrrrrr}
\toprule
$k\sigma$ & Min. gap & Decreases & Max. drop & Final $f_\tau$ & Final residual & Solve (s)\\
\midrule
0 & 0.01000 & 48 & 0.0644 & $-8.1134$ & $4.8\!\times\!10^{-2}$ & 14.4 \\
2 & 0.03000 & 0 & 0.0000 & $-8.0303$ & $5.7\!\times\!10^{-4}$ & 15.4 \\
6 & 0.07000 & 0 & 0.0000 & $-8.0324$ & $1.5\!\times\!10^{-3}$ & 14.6 \\
20 & 0.21000 & 0 & 0.0000 & $-8.0436$ & $6.0\!\times\!10^{-3}$ & 13.6 \\
80 & 0.81002 & 0 & 0.0000 & $-8.2613$ & $4.7\!\times\!10^{-2}$ & 13.4 \\
\bottomrule
\end{tabular}
\end{table}

\begin{table}[!htbp]
\centering
\caption{\textbf{VisDrone budget check on five additional pools per budget.}
Entries give wins/ties/losses of the rounded method against fresh greedy
plus swaps and against the stronger archived baseline. The archived baseline
takes the better of greedy at 12 temperatures and 200 random restarts, all
refined by swaps. $\Delta F_\tau$ is the mean difference against this
stronger baseline. Time is median selection time in seconds.}
\label{tab:locg-budgets}
\small
\begin{tabular}{rllrrr}
\toprule
$k$ & Method & vs. greedy + swaps & vs. archived best & $\Delta F_\tau$ & Select (s)\\
\midrule
15 & Adaptive SCF & 3/1/1 & 0/1/4 & $-0.0014$ & 3.70 \\
15 & SCF + LOCG & 3/2/0 & 0/2/3 & $-0.0030$ & 14.64 \\
20 & Adaptive SCF & 5/0/0 & 5/0/0 & $+0.0488$ & 5.62 \\
20 & SCF + LOCG & 5/0/0 & 5/0/0 & $+0.0515$ & 22.96 \\
25 & Adaptive SCF & 5/0/0 & 3/0/2 & $-0.0008$ & 4.78 \\
25 & SCF + LOCG & 5/0/0 & 2/1/2 & $-0.0002$ & 29.86 \\
\bottomrule
\end{tabular}
\end{table}

\section{AI Use Statement}\label{sec:ai-use}
The authors used generative AI as an assistive tool under their direction
for manuscript redrafting, restructuring, and editing;
literature discovery and suggesting comparisons with relevant prior work; better clarification of claims and mathematical proof checking;
writing and running small numerical checks of mathematical identities and
reported-score arithmetic; \LaTeX{} formatting; assisting with the execution of experiments; and interpretation of
reported results.

\end{document}